\documentclass[journal,10pt,twoside]{IEEEtran}

\ifCLASSINFOpdf
\usepackage[pdftex]{graphicx}
  \graphicspath{{../pdf/}{../jpeg/}}
  \DeclareGraphicsExtensions{.pdf,.jpeg,.png}
\else
  \usepackage[dvips]{graphicx}
  \graphicspath{{../eps/}}
  \DeclareGraphicsExtensions{.eps}
\fi

\usepackage{epstopdf}
\usepackage{amsmath,amsthm, amssymb}
\usepackage{threeparttable}
\usepackage{bm}
\usepackage{multirow}
\usepackage{booktabs}
\usepackage{color}
\usepackage{tabularx}
\usepackage{setspace}
\usepackage{verbatim}
\usepackage{stfloats}
\usepackage{caption}
\usepackage{float}
\usepackage[T1]{fontenc}
\usepackage{soul}
\usepackage{cite}

\usepackage{stfloats}
\usepackage{mathtools}
\usepackage{amsmath}
\usepackage{amsthm}
\usepackage{amssymb}
\usepackage{amsfonts}
\allowdisplaybreaks[0]
\usepackage{subfig}
\usepackage{fancyhdr}
\usepackage{cancel}
\usepackage{units}

\newtheorem{assumption}{Assumption}

\makeatletter
\newif\if@restonecol
\makeatother

\usepackage[linesnumbered,ruled,vlined]{algorithm2e}
\usepackage{algpseudocode}
\usepackage{amsmath}
\usepackage[switch]{lineno}

\begin{document}
	
\title{Distributed Deep Reinforcement Learning Based Gradient Quantization for Federated Learning Enabled Vehicle Edge Computing}

\author{Cui Zhang, Wenjun Zhang, Qiong Wu, ~\IEEEmembership{Senior Member,~IEEE}, Pingyi Fan, ~\IEEEmembership{Senior Member,~IEEE},\\ 
Qiang Fan, Jiangzhou Wang, ~\IEEEmembership{Fellow,~IEEE}, and Khaled B. Letaief, ~\IEEEmembership{Fellow,~IEEE}

\thanks{
This work was supported in part by the National Natural Science Foundation of China under Grant 61701197, in part by the National Key Research and Development Program of China under Grant 2021YFA1000500(4), in part by the Research Grants Council under the Areas of Excellence scheme grant AoE/E-601/22-R, and in part by 111 Project under Grant B23008. (Cui Zhang and Wenjun Zhang contributed equally to this work.)(Corresponding authors: Qiong Wu.)

Cui Zhang is with the School of Internet of Things Engineering, Wuxi Institute of Technology, Wuxi 214122, China (e-mail: faircas85@163.com).

Wenjun Zhang and Qiong Wu are with the School of Internet of Things Engineering, Jiangnan University, Wuxi 214122, China (e-mail: wenjunzhang@stu.jiangnan.edu.cn, qiongwu@jiangnan.edu.cn).

Pingyi Fan is with the Department of Electronic Engineering, Beijing National Research Center for Information Science and Technology, Tsinghua University, Beijing 100084, China (e-mail: fpy@tsinghua.edu.cn).

Qiang Fan is with Qualcomm, San Jose, CA 95110, USA (e-mail: qf9898@gmail.com).

Jiangzhou Wang is with the School of Engineering, University of Kent,
CT2 7NT Canterbury, U.K. (e-mail: j.z.wang@kent.ac.uk).

K. B. Letaief is with the Department of Electrical and Computer Engineering, Hong Kong University of Science and Technology (HKUST), Hong
Kong (e-mail: eekhaled@ust.hk).

Copyright (c) 2024 IEEE. Personal use of this material is permitted. However, permission to use this material for any other purposes must be obtained from the IEEE by sending a request to pubs-permissions@ieee.org.
}}

\maketitle

\begin{abstract}
Federated Learning (FL) can protect the privacy of the vehicles in vehicle edge computing (VEC) to a certain extent through sharing the gradients of vehicles' local models instead of local data. The gradients of vehicles' local models are usually large for the vehicular artificial intelligence (AI) applications, thus transmitting such large gradients would cause large per-round latency. Gradient quantization has been proposed as one effective approach to reduce the per-round latency in FL enabled VEC through compressing gradients and reducing the number of bits, i.e., the quantization level, to transmit gradients. The selection of quantization level and thresholds determines the quantization error, which further affects the model accuracy and training time. To do so, the total training time and quantization error (QE) become two key  metrics for the FL enabled VEC. It is critical to jointly optimize the total training time and QE for the FL enabled VEC. However, the time-varying channel condition causes more challenges to solve this problem. In this paper, we propose a distributed deep reinforcement learning (DRL)-based quantization level allocation scheme to optimize the long-term reward in terms of the total training time and QE. Extensive simulations identify the optimal weighted factors between the total training time and QE, and demonstrate the feasibility and effectiveness of the proposed scheme.

\end{abstract}


\IEEEpeerreviewmaketitle

\section{Introduction}
\label{sec1}
\IEEEPARstart{W}{ith} the rapid development of autonomous driving technology, a large amount of data are generated by various sensors on vehicles, such as cameras, radar, lidar, as well as proximity and temperature sensors. For example, a self-driving car is expected to generate about 1 GB of data per second\cite{wu2024urllc}. Vehicles need powerful computing capability to process and analyze the data to support the modeling training of the vehicular artificial intelligence (AI) applications such as simultaneous localization and mapping (SLAM), augmented reality (AR) navigation, object tracking, and high-definition (HD) map generation\cite{10278180}. However, the computing capability of vehicles is limited. In this situation, vehicle edge computing (VEC) becomes a promising technology to facilitate these applications, where a base station (BS) connected with an edge server can collect and utilize the vehicles' data for model training\cite{long2023power}. However, the raw data generated by a vehicle often contains personal information, thus there may be a risk of data leakage in data privacy in VEC\cite{wu2022mobility}.

Recently, federated learning (FL) has been enabled in VEC\cite{yu2020mobility}. FL can protect the privacy of vehicles to a certain extent through sharing the gradients of vehicles' local models instead of local data\cite{wu2023high}. FL training process consists of multiple rounds. For each round, vehicles individually train their local models using local data, then the BS collects the gradients of local models and aggregates them to obtain a global model, after that the edge server sends the global model to vehicles for the next round training. After meeting the convergence requirement, i.e., reaching the desired precision or the predefined loss value, the entire training process is completed\cite{10506539}. The gradients of vehicles' local models are usually large for the vehicular AI applications, thus transmitting such large gradients would cause large per-round latency.




Gradient quantization may reduce the per-round latency in FL enabled VEC through compressing gradients and reducing the number of bits, i.e., the quantization level, to transmit gradients. In gradient quantization, how to determine the quantization level is a key issue. On one hand, a few quantization level may significantly reduce the transmission overhead, thus cause a low per-round latency and further result in a small total training time. On the other hand, lower number of quantization levels may produce bigger quantization error (QE) due to the heavily compression, which may increase the number of rounds required for convergence and increase the total training time\cite{Oh2022,Alistarh2017}. That means there exists a reasonable gradient quantization level which can adjust the trade off between the total training time and QE for the FL enabled VEC. If only the total training time is optimized, it may lead to the excessive pursuit of large quantization error and the sacrifice of model performance in order to reduce the number of communication rounds, and it is difficult to guarantee the convergence quality and accuracy of the model. Therefore, it is critical to jointly optimize the total training time and QE for the FL enabled VEC. To best of our knowledge, there is no work presented, which motivates us to conduct this work.

In the FL enabled VEC, the channel condition is time-varying due to the time-varying path loss caused by the mobility of the vehicle, which causes challenges to conduct this work\cite{wu2016performance}. Deep reinforcement learning (DRL) is a favorable framework to solve such problem in complex environments\cite{wu2024cooperative}. There are existing papers adopting the centralized DRL framework to dynamically adjust the quantization level in wireless networks\cite{zhu2021decentralized,Wu2022}. However, the centralized DRL framework lead to significantly high traffic load. In addition, for the centralized DRL framework, the high mobility of vehicles makes the BS difficult to collect accurate instantaneous channel state information (CSI). On the contrary, in the distributed DRL framework, each vehicle makes decisions independently and does not need to send all the information to the central node, which can reduce the communication overhead and can reduce the network traffic blocking to the BS at the same time. Since each vehicle makes decisions based on its own state, distributed DRL can better collect the CSI\cite{mills2021multi,wang2021dependent}. Based on the above analysis, it is necessary to find a proper distributed DRL framework to allocate the quantization levels in the FL enabled VEC.

In this paper, we consider the time-varying channel conditions and propose a distributed DRL-based quantization level allocation scheme by optimizing the long-term reward in terms of the total training time and QE\footnote{The source code has been released at: https://github.com/qiongwu86/Distributed-DRL-Based-Gradient-Quantization-for-Federated-Learning-Enabled-VEC}. The main contributions of this paper are summarized as follows.

\begin{itemize}
\item[1)] For each round, if a vehicle is moving at the edge of the BS's coverage, it may have no enough time to participate in the current round of the FL. Hence, the global model may be inaccurate if the vehicle is selected to participate in training of the round. Moreover, the certain vehicles with the similar local model may not need to upload frequently. To avoid such phenomenon, we propose a mobility and model-aware vehicle selection rule to select the vehicles to participate in the training of each round.

\item[2)]  We take the computation time and uploading time in each FL as well as the time-varying channel condition into account and formulate an optimization problem which jointly optimizes the training time and quantization error through allocating quantization level.


\item[3)] We construct the distributed DRL framework including state, action and reward, and adopt the double deep Q-network (DDQN) algorithm to solve the formulated optimization problem, thus the adaptive quantization level allocation can be obtained in a completely distributed way.

\item[4)] Extensive simulation experiments are carried to demonstrated the feasibility and effectiveness of the proposed quantization level allocation scheme.
\end{itemize}

The rest of this paper is organized as follows. Section \ref{sec2} reviews the related works. Section \ref{sec3} introduces the system model. Section \ref{sec4} and Section \ref{sec5} formulate the VEC problem in FL and the optimization problem for quantization level allocation problem, respectively. Section \ref{sec6} presents a solution to the optimized problem based on the distributed DRL. Various simulation results and some discussions are presented in Section \ref{sec7}. Finally, Section \ref{sec8} concludes this paper.

\section{Related Work}
\label{sec2}

In this section, we review the existing works on the optimizations of the training time for FL, the works on the optimizations of the QE for FL, as well as the works on the DRL based performance optimization for FL.

\subsection{Optimizations of the training time for FL}
In \cite{Liu2022}, Liu \emph{et al.} jointly considered data and model and proposed a method to fit the total communication rounds and minimize the total training time of FL through optimizing quantization and bandwidth. In \cite{yang2020delay}, Yang \emph{et al.} proposed a bandwidth allocation optimization method to minimize the total training time of FL based on the relationship between communication rounds and learning accuracy. In \cite{Wang2021}, Wang \emph{et al.} proposed a quantization and bandwidth allocation method to minimize the training time of FL based on the sample of the presence of the outage probability. In \cite{Wan2021}, Wan \emph{et al.} minimized the total training time of FL in wireless networks through adaptive resource scheduling. In \cite{Chen2021}, Chen \emph{et al.} minimized the total training time of FL by combining the optimization of user selection and resource block (RB) allocation. In \cite{jiang2022}, Jiang \emph{et al.} proposed a novel FL method to adapt to the model size, reducing communication and computation overhead, minimizing the overall training time while maintaining a similar accuracy to the original model. In \cite{Nguyen}, Nguyen \emph{et al.} proposed an efficient FL algorithm to allow a small number of clients to participate in the training process based on the unbiased sampling strategy in each round, and then adopted the proposed FL algorithm to formulate the wireless internet of things (IoT) resource allocation problem to minimize the training time. However, these works have not optimized the quantization error of FL.

\subsection{Optimizations of the quantization error for FL}
In \cite{Wang2022}, Wang \emph{et al.} minimized quantization error of FL through jointly allocating wireless bandwidth and quantization level. In \cite{Honig2022}, Hönig \emph{et al.} minimized the communication cost under the constraint of quantization error and kept the quantization error of FL at the static quantization level by assigning different quantization levels to the client. In \cite{Jhunjhunwala2021}, Jhunjhunwala \emph{et al.} proposed an adaptive quantization strategy to achieve a low quantization error upper bound of FL by changing the quantization level during training. In \cite{Shlezinger,Shlezinger2021}, Shlezinger \emph{et al.} proposed a general vector quantization scheme based on dithered lattice quantization to reduce the quantization error of FL. In \cite{ChenShengbo}, Chen \emph{et al.} proposed the federated learning with heterogeneous quantization (FEDHQ) by assigning aggregate weights to minimize an upper bound on the convergence rate, which is a function of quantization error. In \cite{Nori2021}, Nori \emph{et al.} minimized the quantization error of FL for a given wall-clock time by dynamically adjusting the local update coefficients. However, these works have not considered optimizing the training time of FL.

\subsection{DRL based optimizations for FL}

In \cite{WANG2022joint}, Wang \emph{et al.} proposed a method based on the multi-agent deep deterministic policy gradient (MADDPG) to realize fully distributed self-allocation of multi-dimensional resources and optimize the cost of participants in FL. In \cite{Zhang2021}, Zhang \emph{et al.} proposed an FL algorithm assisted by DRL, which can balance the privacy and efficiency of data training for the industrial internet of things (IIoT) devices. In \cite{Do2022}, Do \emph{et al.} proposed a DRL-based framework for joint UAV placement and resource allocation to enable the sustainable FL with energy harvesting user devices. In \cite{Wu2022fed}, Wu \emph{et al.} proposed a DRL-based optimization and clustering approach for computation constrained devices to accelerate local training of FL by offloading deep neural networks (DNNs) to the server. However, these works have not considered jointly optimizing the training time and quantization error of FL.

As mentioned above, there is no work jointly optimizing the training time and quantization error of FL in VEC based on DRL, which is the motivation of our work in this paper.

\section{System Model}
\label{sec3}

\begin{table*}
	\footnotesize
	\caption{The summary for notations.}
	\label{tab1}
	\centering
	\begin{tabular}{|p{2.1cm}<{\centering}|p{5.8cm}|p{2cm}<{\centering}|p{5.8cm}|}
		\hline
		\textbf{Notation} &\textbf{Description} &\textbf{Notation} &\textbf{Description}\\
		\hline
		$N$ &the number of vehicles covered by a BS& $\mathcal{V}$ &a set of vehicles covered by a BS\\
		\hline
		$\mathcal{D}_n$ &the training dataset of vehicle $V_n$ &$D$ &the size of the training dataset\\
		\hline
		$r$ &federated learning round &$\mathcal{K}^r$ &the set of selected vehicles in round $r$\\
		\hline
		$K$ &the number of selected vehicles &$\mathbf{w}_g^r$ & the global
		model in round $r$\\
		\hline
		$d$ &the model size &$\mathbf{g}_k^r$ &the local stochastic gradient in round $r$\\
		\hline
		$\mathcal{Q}(\mathbf{g}_k^r)$ &the quantized gradient &$x_n^r$ &the X-axis coordinate of $V_n$\\
		\hline
		$v_n^r$ &the velocity of $V_n$ &$T_g^r$ &the average time of a round \\
		\hline
		$\varphi^{\ast}$ &the decision threshold &$\mathbf{w}_n^r$ &the local model of $V_n$\\
		\hline
		$R_B$ &the radius of the BS &$T_{n}^{r,res}$ &\multicolumn{1}{m{6cm}|}{the residence time of $V_n$ within the coverage of the BS}\\
		\hline
		$\varphi_{n}^r$ &the utility of $V_n$  &$\alpha_{n}^r$ &the model similarity\\
		\hline
		$\beta_{n}^r$ &the estimated
		leaving time &$\xi _n^r$ &the decision result of the vehicles\\
		\hline
		$F(\mathbf{w}_k^r)$ & the local loss function &$f(\mathbf{w}_g^{r};\mathbf{x}_i,y_i)$ & the sample loss function\\
		\hline
		$g_k^{r,j}$& each element $j$ of the local gradient &$\Vert\mathbf{g}_k^{r}\Vert$ & the vector norm\\
		\hline
		$sgn(g_k^{r,j})$& the sign of each element &$\xi_{j}(\mathbf{g}_k^{r},q_k^{r})$ &the normalized quantization value of each element \\
		\hline
		$q_k^r$ & the number of quantization levels &$\lambda$ &the loss optimality difference\\
		\hline
		$\mathbf{w}_{\ast}$ &the optimal global model &$T_{k}^{r,\text{fed}}$ &the federated learning latency for one round of $V_k$\\
		\hline
		$T_{k}^{r,\text{comp}}$ &the local model training computation time & $T_{k}^{r,\text{upload}}$ &the local parameter upload delay\\
		\hline
		$T_u$& the upper limit of FL delay for a round&$R_{\lambda}$ &\multicolumn{1}{m{6cm}|}{the minimum number of rounds to achieve the $\lambda$-optimal difference}\\
		\hline
		$T_k^{\text{total}}$ &\multicolumn{1}{m{6cm}|}{the requirement of total training time over $R_\lambda$ rounds} &$f_k$ & the CPU frequency of $V_k$\\
		\hline
		$c$ &\multicolumn{1}{m{6cm}|}{the number of processing cycles to execute a batch of samples} & $\zeta_k^{r}$ &is the number of uploading bits\\
		\hline
		$R_k$ & the transmission rate of $V_k$ &$\gamma_{k}$ &SNR of $V_k$ \\
		\hline
		$d_{k}$ &\multicolumn{1}{m{6cm}|}{the distance between $V_k$ and the BS} &$\alpha$ &the path loss exponent\\
		\hline
		$p_k$ &the transmit power of vehicle $V_k$ &$h_{k}$ &the channel gain between $V_k$ and the BS\\
		\hline
		$\sigma^2$ &the noise power &$s_{k,t}$ &the state of $V_k$ at time step $t$\\
		\hline
		$a_{k,t}$ & the action of $V_k$ at time step $t$ &$r_{k,t}$ &the reward function of $V_k$ at time step $t$\\
		\hline
		$w_1,w_2$ &the weighted factors of reward &$\gamma$ &discounting factor of long-term reward\\
		\hline
		$T$ &the upper limit of time step index &$G(\pi_k)$ &long-term discounted reward of $V_k$\\
		\hline
		$\pi_k^{\ast}$ &the optimal policy &$Q(s_t,a_t;\theta)$ & prediction value\\
		\hline
		$y_t$ & target value &$\mathcal{B}$ &replay buffer\\
		\hline
		$I$ & size of mini-batch &$C$ &the period of the target network update\\
		\hline
		$\delta_t$ &the time difference error &$T_I$ &the time step interval\\
		\hline
		$\epsilon$ & exploration probability  &$L(\theta)$ &the loss function of prediction network\\
		\hline
		$\nabla L(\theta)$ &the gradient of the loss function &$\theta$ &parameter of prediction network \\
		\hline
		$\theta^{'}$ &parameter of target network &$ $ &\\
		\hline
	\end{tabular}
\end{table*}

\begin{figure*}
	\centering
	\includegraphics[width=5.5in]{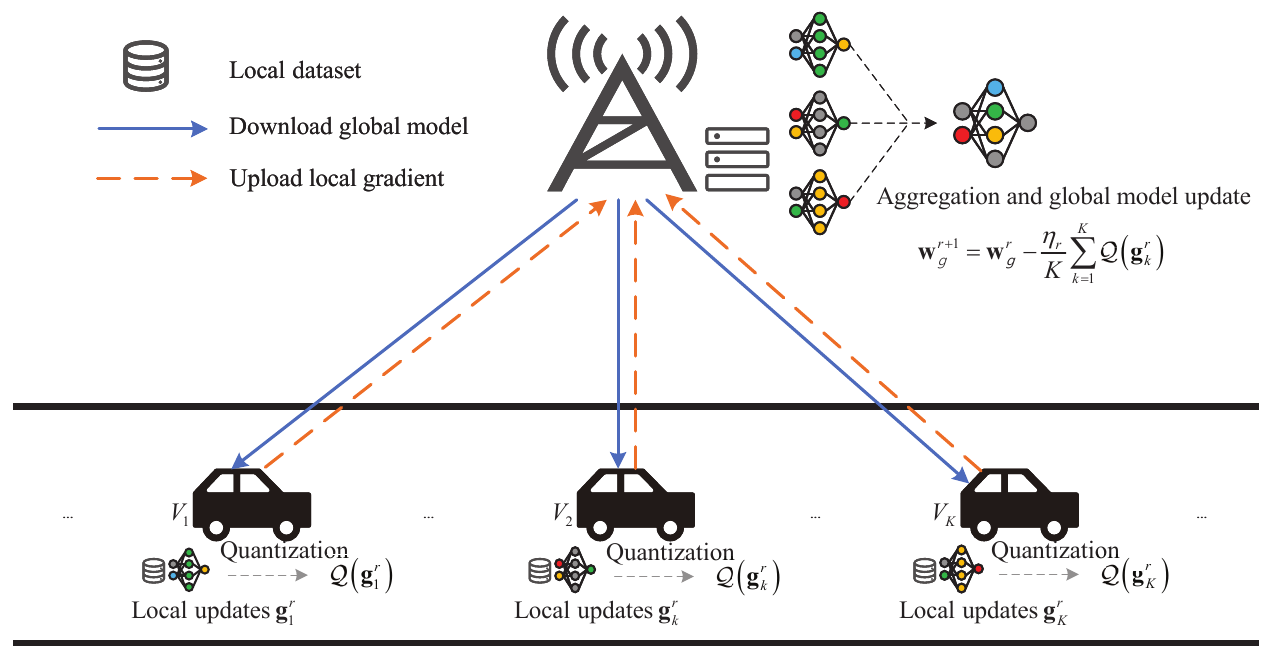}
	\caption{System model.}
	\label{fig1}
\end{figure*}


We consider a quantized FL system consisting of $N$ vehicles and a BS connected with an edge server. The vehicles are driving in the coverage of the BS. The velocities of different vehicles follow Gaussian distribution and the velocity of each vehicle is not changed. The driving duration of the vehicles is divided into time slots\cite{wu2014performance1}. The set of the vehicles is denoted by $\mathcal{V}=\{V_1,V_2,V_3,\vcenter{\hbox{$\cdots$}},V_n, \vcenter{\hbox{$\cdots$}}, V_N\}$. Each vehicle $V_n$ is equipped with a MEC processor that provides the local processing capability. Vehicles get the data samples through sensors or communication, and each vehicle $V_n$ holds the local dataset $\mathcal{D}_n$ with a uniform size of $D$, the dataset $\mathcal{D}_n$ contains data samples $\left\{(\mathbf{x}_i,y_i)\right\}_{i=1}^{D}$, where $\mathbf{x}_i$ represents the feature of the data sample and $y_i$ represents the label of the data sample. The features of the data sample are used as the input of the model training, and the labels of the data sample are the expected outputs corresponding to the input data, which are used to train and validate the model\cite{wu2015performance,wu2014performance2}.

Vehicles adopt the quantized FL to obtain the global model through training in rounds. The BS first selects vehicles based on the mobility and model-aware vehicle selection algorithm to participate in training at the beginning of each round $r$. Assume that the set of selected vehicles in each round $r$ is denoted by $\mathcal{K}^{r}=\{V_1,V_2,V_3,\vcenter{\hbox{$\cdots$}},V_k,\vcenter{\hbox{$\cdots$}},V_{K}\}$, where $K$ is the number of the selected vehicles in each round $r$ ($K$ $\leq$ $N$). Then each selected vehicle $V_k$ downloads the current global model $\mathbf{w}_g^{r} \in \mathbb{R}^d$  from the BS, where $d$ denotes the model size, then each vehicle randomly selects samples from the dataset $\mathcal{D}_k$ in a uniform manner to obtain a mini-batch and computes the local stochastic gradient $\mathbf{g}_k^{r}$ based on the mini-batch. Next, each selected vehicle makes a decision based on its local observation to determine the quantization level which can optimize the training time and quantization error, then adopts the widely-used stochastic gradient quantization to quantize the local gradient and obtain the quantized gradient $\mathcal{Q}(\mathbf{g}_k^{r})$\cite{Alistarh2017}. Then, all selected vehicles upload their quantized gradients to the BS, where the orthogonal frequency division multiple access (OFDMA) technology\cite{Zhu2009,Zhu2012} is adopted for the communication between each vehicle and the BS. Afterwards, the BS aggregates the quantized gradients of all vehicles to obtain the updated global model $\mathbf{w}_g^{r+1}$ which is used for the training of the next round. The round is repeated until the global model is convergent. The network model is shown in Fig. \ref{fig1}. For ease of understanding, the key notations have been summarized in TABLE \ref{tab1}.
\section{Quantized Federated Learning Algorithm for Vehicle Edge Computing}
\label{sec4}
In this section, we introduce the quantized FL algorithm for VEC. The pseudocode of the algorithm is shown in Algorithm 1.


\begin{algorithm}
\caption{Quantized Federated Learning Algorithm.}
\label{al1}
\KwIn{vehicles $\mathcal{V}$}
\KwOut{the optimal global model $\mathbf{w}_\ast$}
\textbf{Initialize} the global model $\mathbf{w}_g^0$\\
\For{each round $r= 0$ to $R$}
{
	Each vehicles $V_n$ downloads the global model $\mathbf{w}_g^{r}$\\
	\For{each vehicle $V_n$}
	{
	Obtain mobility information $x_n^{r}$, $v_n^{r}$\\
	Receive area information $T_{g}^{r}$, $\varphi^{\ast}$\\
	Load the vehicle model parameter $\mathbf{w}_n^{r}$\\
	Estimate residence time $T_{n}^{r,res}$ according to Eq. (\refeq{eq:T_residence})\\
	Calculate utility $\varphi_{n}^{r}$ according to Eq. (\refeq{eq:phi_n})\\
	Calculate necessity $\xi _n^{r}$ according to Eq. (\refeq{eq:xi_n})\\
	\lIf{$\varphi_{n}^{r}\geq \varphi^{\ast}$}{$\xi _n^{r}=1$}
	\lElse{$\xi _n^{r}=0$}
	Obtain the set of selected vehicles ${\mathcal{K}^{r}=\{V_n \textbar \xi_n^{r} = 1,n \in \mathcal{N}\}}$
	}
	\For{each vehicle $V_k$ in $\mathcal{K}^{r}$}
	{
	Calculates the local loss function $F(\mathbf{w}_k^{r})$ according to Eq. (\refeq{eq:F_wn})\\
	Calculates the local gradient $\mathbf{g}_k^{r}$ according to Eq. (\refeq{eq:gradient})\\
	Quantize the local gradient according to Eq. (\refeq{eq:gradient1})\\

	}
Each vehicle transmit the quantized local gradient $\mathcal{Q}(\mathbf{g}_k^{r})$ and its local loss function $F(\mathbf{w}_k^{r})$ to the BS\\
BS aggregates the quantized local gradients and updates the global model according to Eq. (\refeq{eq:w_gr})\\
BS calculates the global loss function $F(\mathbf{w}_g^{r})$ according to Eq. (\refeq{eq:F_wgr})\\
\While{$F(\mathbf{w}_g^{r})-F(\mathbf{w}_{\ast}^{r}) \leq \lambda$}
{The algorithm stops}
}
\end{algorithm}	
First, the BS initializes the global model for round $0$, which is denoted by $\mathbf{w}_g^{0}$. Then the training is conducted for $R+1$ rounds. For each round $r$, each vehicle $V_n$ first downloads the global model $\mathbf{w}_g^{r}$ from the BS (lines 1-3). Then, the mobility and model-aware vehicle selection algorithm is conducted to select the vehicles which have enough time to participate in this round and do not have similar models.

Specifically, a planar rectangular coordinate system is established with the BS as the center and each vehicle $V_n$ first obtains its own mobility information, i.e., the current X-axis coordinate of $V_n$, denoted by $x_n^{r}$, and the velocity of $V_n$, denoted by $v_n^{r}$. Then, the BS calculates the average time of a round, denoted by $T_g^{r}$, which is calculated by averaging the durations of the rounds before round $r$ and sends $T_g^{r}$ as well as the decision threshold $\varphi^{\ast}$ to each vehicle. After that each vehicle $V_n$ loads its local model $\mathbf{w}_n^{r}$ (lines 4-7).

Assuming that the radius of the BS is $R_B$, each vehicle $V_n$ calculates its residence time within the coverage of the BS as (line 8)
\begin{equation}
T_{n}^{r,res}=\frac{R_B-x_n^{r}}{v_n^{r}}.
\label{eq:T_residence}
\end{equation}

Each vehicle $V_n$ calculates the utility $\varphi_{n}^{r}$ to determine if vehicle $V_n$ is selected to participate in round $r$. Vehicles with similar local models to the global model contribute less to the training of the current global model and do not need to participate in the training frequently. Meanwhile, the selected vehicles need to have enough time to participate in the current round of training and upload the local model. Thus, $\varphi_{n}^{r}$ is affected by both model similarity $\alpha_{n}^{r}$ and vehicle's residence time in the coverage $\beta_{n}^{r}$, i.e., (line 9)
\begin{equation}
\varphi_{n}^{r}=\alpha_{n}^{r}+\beta_{n}^{r},
\label{eq:phi_n}
\end{equation}
where
\begin{equation}
\alpha_{n}^{r}=\frac{\Vert \mathbf{w}_{n}^{r}-\mathbf{w}_{g}^{r}\Vert}{\text{max}\{\Vert \mathbf{w}_{n}^{r}\Vert,\Vert \mathbf{w}_{g}^{r}\Vert\}},
\label{eq:alpha_n}
\end{equation}
\begin{equation}
\beta_{n}^{r}=\frac{ T_{n}^{r,res}-T_{g}^{r}}{\text{max}\{ T_{n}^{r,res}, T_{g}^{r}\}},
\label{eq:beta_n}
\end{equation}
here $\alpha_{n}^{r}$ is used to measure the similarity between the local model of the vehicle and the global model, and $\beta_{n}^{r}$ quantifies the relationship between the expected residence time of the vehicle within the coverage of the BS and the average time of a round. Both $\alpha_{n}^{r}$ and $\beta_{n}^{r}$ are normalized metrics, which means that they have the same degree of influence on $\varphi_{n}^{r}$, $\alpha_{n}^{r} \in [0,1]$, $\beta_{n}^{r} \in [-1,1)$.

Then, each vehicle $V_n$ calculates $\xi_{n}^{r}$ which is a binary indicator to show if it is selected (lines 10-12), i.e.,
\begin{equation}
{\xi_n^{r}}= \left\{ {\begin{array}{*{20}{c}}
1,&{\text{if}\;\varphi_{n}^{r}\geq \varphi^{\ast}},\\
0,&{{\text{else.}}}
\end{array}} \right.
\label{eq:xi_n}
\end{equation}

Up to now, each vehicle $V_n$ knows whether it is selected, i.e., the set of the selected vehicles ${\mathcal{K}^{r}=\{V_1,V_2,V_3,\dots,V_k,\dots,V_{K}\}=\{V_n \textbar \xi_n^{r} = 1,n \in \mathcal{N}\}}$ is obtained, where $\xi_n^{r} = 1$ indicates that vehicle $V_n$ is selected, otherwise $\xi_n^{r} = 0$ (line 13).

After that each selected vehicle $V_k$ calculates the local loss function $F(\mathbf{w}_k^{r})$ as (lines 14-15)
\begin{equation}
	F(\mathbf{w}_k^{r})=\frac{1}{D}\sum_{(\mathbf{x}_i,y_i) \in D_k}f(\mathbf{w}_g^{r};\mathbf{x}_i,y_i),
	\label{eq:F_wn}
\end{equation}
where $f(\mathbf{w}_g^{r};\mathbf{x}_i,y_i)$ denotes the sample loss function which is specified by the learning task for $V_k$ and quantifies the training loss of global model $\mathbf{w}_g^{r}$ on the training data $\mathbf{x}_i$ and its label $y_i$.

Then each selected vehicle $V_k$ calculates the local gradient $\mathbf{g}_k^{r}$ using the local dataset $\mathcal{D}_k$, and we have (line 16)
\begin{equation}
	\mathbf{g}_k^{r}= \nabla F(\mathbf{w}_k^{r}).
	\label{eq:gradient}
\end{equation}


Next, each selected vehicle $V_k$ allocates its quantization level for round $r$, denoted as $q_k^{r}$,  and uses a widely-used stochastic gradient quantization to quantize the local gradient \cite{Alistarh2017}. For vehicle $V_k$, each element $j$ of its local gradient $\mathbf{g}_k^{r}$, denoted by $g_k^{r,j}$, is quantized as (line 17)
\begin{equation}
	\begin{aligned}
		{\cal Q}\left( {{g_k^{r,j}}} \right) = &\left\| {\mathbf{g}_k^{r}} \right\| \cdot {\mathop{\rm sgn}} \left( {{g_k^{r,j}}} \right) \cdot {\xi _j}\left( {{\mathbf{g}_k^{r}},q_k^{r}} \right),\;\\
		&\forall j \in \{1,2,\cdots,d\}, {\cal Q}\left( {{g_k^{r,j}}} \right) \in \mathbb{R}^d,
		\label{eq:gradient1}
	\end{aligned}
\end{equation}
where the quantization function ${\cal Q}\left( {{g_k^{r,j}}} \right)$ consists of three parts, i.e., the vector norm $\Vert\mathbf{g}_k^{r}\Vert$, the sign of each element $\mathrm{sgn}(g_{k}^{r,j})$ , and the normalized quantization value of each element $\xi_{j}(\mathbf{g}_k^{r},q_k^{r})$, which is a random variable defined as
\begin{equation}
	{\xi_j}\left( \mathbf{g}_k^{r},q_k^{r} \right) = \left\{ {\begin{array}{*{20}{c}}
			(l+1)/q_k^{r},&{\text{w.p.}\;\frac{{\left| {{g_k^{r,j}}} \right|}}{{\left\| {\mathbf{g}_k^{r}} \right\|}}q_k^{r} - l},\\
			l/q_k^{r},&{{\text{otherwise.}}}
	\end{array}} \right.
\end{equation}
Here, $0\leq l < q_k^{r}$ is an integer such that $\frac{g_k^{r,j}}{\left\| {\mathbf{g}_k^{r}} \right\|} \in \left[ \frac{l}{q_k^{r}},\frac{l+1}{q_k^{r}}\right)$.

Next, each selected vehicle $V_k$ transmits the quantized local gradient , i.e., $\mathcal{Q}(\mathbf{g}_k^{r})$, and its local loss function $F(\mathbf{w}_k^{r})$ to the BS. After the BS receives the quantized local gradients and local loss functions from all vehicles, the BS aggregates the quantized local gradients and updates the global model as (lines 18-19)
\begin{equation}
	\mathbf{w}_g^{r+1} = \mathbf{w}_g^{r}-\frac{\eta_r}{K}\sum^{K}_{k=1}\mathcal{Q}(\mathbf{g}_k^{r}).
	\label{eq:w_gr}
\end{equation}

Then the BS calculates the global loss function as (line 20)
\begin{equation}
	F(\mathbf{w}_g^{r})= \frac{1}{K}\sum^{K}_{k=1}F(\mathbf{w}_k^{r}).
	\label{eq:F_wgr}
\end{equation}

Then the global model that minimizes the global loss function from the first round to the current round $r$ is calculated as,
\begin{equation}
\mathbf{w}_{\ast}^{r}=\arg\mathop{\min}_{\mathbf{w}_g^{r}} F(\mathbf{w}_g^{r}).
	\label{eq1}
\end{equation}

 Next the BS calculates the difference between the global loss of the current round, i.e., $F(\mathbf{w}_g^{r})$, and the global loss of the optimal global model, i.e., $F(\mathbf{w}_{\ast}^{r})$. If $\lambda$-optimal difference is satisfied, i.e.,
 \begin{equation}
 F(\mathbf{w}_g^{r})-F(\mathbf{w}_{\ast}^{r}) \leq \lambda,
 	\label{convergence}
\end{equation}
the global model is converged, where $\lambda$ is the loss optimality difference. In this case, the training process stops and the algorithm outputs the global model $\mathbf{w}_g^{r}$ as the optimal global model $\mathbf{w}_{\ast}$ (lines 21-22). Otherwise, all vehicles download the updated global model from the BS to start the next round. When the number of rounds reaches $R$, the training process stops and the algorithm outputs the global model $\mathbf{w}_{\ast}^{R}$ as the optimal global model $\mathbf{w}_{\ast}$.

\section{Problem Formulation}
\label{sec5}

In this section, we formulate the problem to jointly optimize the training time and QE. We first formulate the training time and QE, respectively. Then we formulate the optimization problem.

\subsection{Training time}

The FL delay of round $r$ for vehicle $V_k$, denoted by $T_{k}^{r,\text{fed}}$, consists of four components: the computation time $T_{k}^{r,\text{comp}}$ for local model training; uploading time $T_{k}^{r,\text{upload}}$; global aggregation delay; downloading time of the global model. Similar to \cite{WANG2022joint}, since the aggregation in the BS is performed extremely fast and the downlink bandwidth is large enough, we assume that the global aggregation delay and downloading time of the global model can be ignored, thus $T_{k}^{r,\text{fed}}$ is defined as
\begin{equation}
	T_{k}^{r,\text{fed}}=T_{k}^{r,\text{comp}}+T_{k}^{r,\text{upload}}.
	\label{eq17}
\end{equation}

We define $R_\lambda^r$ as the estimated minimum convergence round in round $r$ to achieve the convergence of global model. We estimate $T_{k}^{r,\text{fed}}$ as the training time of each round before aggregation, thus for round $r$ the requirement of the total training time over $R_\lambda^r$ is estimated as
\begin{equation}
	\label{eq:total-time}
	T_k^{r,\text{total}} = R_{\lambda}^r \cdot T_{k}^{r,\text{fed}},
\end{equation}


Next, the computation time $T_{k}^{r,\text{comp}}$, uploading time  $T_{k}^{r,\text{upload}}$ and minimum convergence round $R_{\lambda}^r$ are formulated, respectively.

\subsubsection{Computation time}

Let $c$ denote the number of processing cycles for a particular vehicle to update the local model based on a batch of samples, and $f_k$ denotes the CPU frequency of $V_k$. Therefore, the computation time of round $r$ is given by \cite{Ren2020}
\begin{equation}
	T_{k}^{r,\text{comp}}=\frac{c}{f_{k}}.
	\label{eq17}
\end{equation}

\subsubsection{Uploading time}

The latency for $V_{k}$ to upload the quantized local gradient of round $r$ is
\begin{equation}
	T_{k}^{r,\text{upload}}=\frac{\zeta_k^{r}}{R_{k}^r},
\end{equation}
where $\zeta_k^{r}$ is the number of uploading bits after stochastic quantization and $R_k^r$ is the transmission rate of $V_k$ for round $r$.

For quantizing any element $g_k^{r,j}$ in the local gradient $\mathbf{g}_k^{r}$ according to Eq. (\ref{eq:gradient1}), we need to encode  $\Vert\mathbf{g}_k^{r}\Vert$, $\mathrm{sgn}(g_{k}^{r,j})$ and $\xi_{j}(\mathbf{g}_k^{r},q_k^{r})$ into bits. Particularly, $\mathrm{sgn}(g_{k}^{r,j})$ is encoded by utilizing one bit. Assume that $\xi_{j}(\mathbf{g}_k^{r},q_k^{r})$ follows uniform distribution\footnote{In practice, $\xi_{j}(\mathbf{g}_k^{r},q_k^{r})$ may follow non-uniform distribution. As reported in \cite{Alistarh2017}, such non-uniform distribution is a more efficient coding method compared with uniform distribution. However, the non-uniform distribution is much more complicated to analyze and optimize, hence the uniform distribution is considered in this paper and we will consider the non-uniform distribution for future work.}. According to \cite{Cover2006}, it takes at least $\log_2(1+q_k^{r})$ bits to encode each $\xi_{j}(\mathbf{g}_k^{r},q_k^{r})$. Moreover, similar to \cite{Shlezinger}, the bits to encode $\Vert\mathbf{g}_k^{r}\Vert$ is typically negligible for large models. Since $\mathbf{g}_k^{r}$ contains $d$ elements, according to the above analysis $\zeta_k^{r}$ can be approximated as follows
\begin{equation}
	\zeta_k^{r} = [1+\log_{2}(q_k^{r}+1)]d.
	\label{eq:S}
\end{equation}

Next, the transmission rate of $V_k$, denoted by $R_k^r$, is formulated. OFDMA is adopted for wireless transmission between the BS and vehicles. The bandwidth is $B$ and can be divided into $W$ orthogonal subcarriers. Each selected vehicle is assigned with an orthogonal subcarrier. The mobility of each selected vehicle $V_k$ has an impact on the transmission rate $R_k^r$, which depends on the distance $d_k$ between $V_k$ and the BS \cite{Liu2022mobility,Luo2021}. According to Shannon's theorem, the achievable transmission rate of vehicle $V_k$ can be represented as
\begin{equation}
	R_k^r=\frac{B}{W}\log_{2}(1+\gamma_k^r),
	\label{eq:Rn}
\end{equation}

\begin{equation}
	\gamma_k^r=\frac{p_kh_k^r(d_k^r)^{-\alpha}}{\sigma^2},
	\label{eq:SINR_n}
\end{equation}
where $\gamma_k^r$ is the signal-to-noise ratio (SNR) of vehicle $V_k$, $p_k$ is the transmit power of vehicle $V_k$, which remains constant, $h_k^r$ and $d_k^r$ are the channel gain and distance between vehicle $V_k$ and the BS, respectively, where $h_k^r$ is calculated according to the normal distribution in \cite{zhu2021decentralized}, $\alpha$ is the path loss exponent and $\sigma^2$ is the noise power, $d_k^r$ can be calculated as
\begin{equation}
	d_k^r=\sqrt{(x_k^r-x_B)^2+(y_k-y_B)^2},
	\label{eq:d_n}
\end{equation}
where $(x_B, y_B)$ is the coordinate of the BS, and $(x_k^r, y_k)$ is the coordinate of $V_k$.


\subsubsection{Minimum convergence round}
The minimum convergence round is derived by analyzing the convergence of the quantized federated learning. To facility derivation, we make some assumptions on the local loss function of $V_k$ for round $r$, denoted by $F(\mathbf{w}_k^{r})$, as follows:
\begin{assumption}[\emph{L}-Smoothness]\label{asp:smoothness}
	The local loss function $F(\mathbf{w}_k^{r})$ is $L$-smooth for $\forall k \in \mathcal{K}^{r}$ and $\forall r$, i.e., $F(\mathbf{w}_{k,i}^{r}) \leq F(\mathbf{w}_{k,j}^{r}) + \left(\mathbf{w}_{k,i}^{r} - \mathbf{w}_{k,j}^{r}\right)^{T}\nabla F(\mathbf{w}_{k,j}^{r}) + \frac{L}{2}\left\Vert \mathbf{w}_{k,i}^{r} - \mathbf{w}_{k,j}^{r} \right\Vert^{2}$, $\forall \mathbf{w}_{k,i}^{r}$ and $\mathbf{w}_{k,j}^{r} \in \mathbb{R}^d$ are the $i$-th and $j$-th local model of $V_k$ for round $r$.
\end{assumption}

\begin{assumption}[$\mu$-Strongly convexity]\label{asp:convexity}
	The local loss function $F(\mathbf{w}_k^{r})$ is $\mu$-strongly convex for $\forall k \in \mathcal{K}^{r}$ and $\forall r$, i.e., $F(\mathbf{w}_{k,i}^{r}) \ge F(\mathbf{w}_{k,j}^{r}) + \left(\mathbf{w}_{k,i}^{r} - \mathbf{w}_{k,j}^{r}\right) + \left(\mathbf{w}_{k,i}^{r} - \mathbf{w}_{k,j}^{r}\right)^{T}\nabla F(\mathbf{w}_{k,j}^{r}) + \frac{\mu}{2}\left\Vert \mathbf{w}_{k,i}^{r} - \mathbf{w}_{k,j}^{r} \right\Vert^{2}$.
\end{assumption}

\begin{assumption}[Boundedness of variance]\label{asp:bounded}
	The variance of stochastic gradient $\mathbf{g}_{k}^{r}$ of the local loss function $F(\mathbf{w}_k^{r})$ for $\forall k \in \mathcal{K}^{r}$ and $\forall r$, satisfy that:
	$\mathbb{E}[\Vert\mathbf{g}_{k}^{r} - \nabla F(\mathbf{w}_k^{r})\Vert^{2}] \leq \delta_{k}^{2}$.
\end{assumption}

\begin{assumption}[Unbiased]\label{asp:unbiased}
	The mean of stochastic gradient $\mathbf{g}_{k}^{r}$ of the local loss function $F(\mathbf{w}_k^{r})$ for $\forall k \in \mathcal{K}^{r}$ and $\forall r$, satisfy that:
	$\mathbb{E}[\mathbf{g}_{k}^{r}] = \nabla F(\mathbf{w}_k^{r})$.
\end{assumption}

Note that assumptions \ref{asp:smoothness} and \ref{asp:convexity} on the local loss function can be satisfied by many typical learning models, such as logistic regression, linear regression, and softmax classifiers. Assumptions 3 and 4 are general. Based on Assumptions 1-4, the optimality gap between the average loss function of the global model for a round $r'$ after round $r$, denoted by $\mathbb{E}\left[F(\mathbf{w}_g^{r'})\right]$ $(r'>r)$, and the loss function of optimal global model for round $r$, i.e., $F(\mathbf{w}_{\ast}^{r})$, is upper bounded by \cite{Liu2022}:

\begin{equation}
	\begin{aligned}
		&\mathbb{E}\left[F(\mathbf{w}_g^{r'})\right]-F(\mathbf{w}_{\ast}^{r})\\
		&\leq \frac{\alpha \kappa}{r'+2\alpha \kappa-1}\left(L\left\Vert \mathbf{w}_g^{0} - \mathbf{w}_{\ast}^{r} \right\Vert^2 + \frac{2\Gamma}{\mu}\right),
	\end{aligned}
\end{equation}
where $\alpha = \frac{\sqrt{d}}{q_k^{r}K}+1$, $\kappa = \frac{L}{\mu}$, $\Gamma = 2 LF_{\delta} +\frac{1}{K}\sum_{k=1}^{K}\delta_{k}^{2}$, here $F_{\delta} = F(\mathbf{w}_{\ast}^{r}) - \frac{1}{K}\sum_{k=1}^{K}F_{k}^{\ast}$ with $F_{k}^{\ast}=\min\limits_{\mathbf{w}_k^{r}}F(\mathbf{w}_k^{r})$, $\mathbf{w}_g^{0}$ is the initial global model of the training process. By letting the upper bound satisfy the convergence constraint\footnote{We note that optimization using upper bound as an approximation has also been adopted in \cite{Wang2019} and resource allocation based literature \cite{Chen2021}\cite{Yang2020}.}, we can obtain
\begin{equation}\label{eq:T_epsilon1}
	\frac{\alpha \kappa}{R_{\lambda}^r+2\alpha \kappa-1}\left(L\left\Vert \mathbf{w}_g^{0} - \mathbf{w}_{\ast}^{r} \right\Vert^2 + \frac{2\Gamma}{\mu}\right) \leq \lambda.
\end{equation}

Letting the left hand equals to the right hand in Eq. \eqref{eq:T_epsilon1}, when $\lambda$ is given, we can derive $R_{\lambda}$ based on the fact that the minimum convergence round should be integer, i.e.,
\begin{equation}\label{eq:T_epsilon}
	R_{\lambda}^r = \left\lceil \left(\frac{\sqrt{d}}{q_k^{r}K}+1\right)\left(\frac{L{\Vert \mathbf{w}_g^{0} - \mathbf{w}_{\ast}^{r} \Vert}^{2}+\frac{2 \Gamma}{\mu}}{\lambda} -2\right)\kappa+1\right\rceil.
\end{equation}

\subsection{Quantization error}
According to \cite{Alistarh2017}, the random quantization function $\mathcal{Q}(\mathbf{g}_k^{r})$ is unbiased, i.e., $\mathbb{E}[\mathcal{Q}(\mathbf{g}_k^{r})]=\mathbf{g}_k^{r}$ for any given $\mathbf{g}_k^{r}$. Moreover, assume that $d \ge ({q_k^{r}})^2$, the quantization function is constrained by a upper bound, i.e., $\mathbb{E}[\left\Vert\mathcal{Q}(\mathbf{g}_k^{r})-\mathbf{g}_k^{r}\right\Vert]^{2} \leq \frac{\sqrt{d}}{q_k^{r}}\left\Vert\mathbf{g}_k^{r}\right\Vert^{2}$. Similar with  \cite{Wang2021}, we estimate the upper bound as the quantization error of vehicle $V_k$ in round $r$, i.e.,
\begin{equation}\label{eq:E_n}
E_k^{r} = \frac{\sqrt{d}}{q_k^{r}}\Vert \textbf{g}_k^{r} \Vert^{2}.
\end{equation}

According to Eq. (\ref{eq:S}) and Eq. (\ref{eq:E_n}), we can get the following conclusion, i.e., a higher quantization level would lead to more bit to transmit and a less quantization error $E_k^{r}$.
\subsection{Optimization problem}

Our optimization objective is to allocate quantization levels to jointly optimize the estimated training time and quantization error for each vehicle $V_k$ in each round $r$, which can be translated into the following problem:
\begin{align}
	\label{eq:problem}\textbf{P1:}\;\;\;\underset{q_k^r}{\text{min   }}
	\;\;\;&\omega_{1}\cdot T_k^{r,\text{total}}+\omega_{2}\cdot E_k^r,\\  \text{s.t.} \;\;\;
	&\nonumber q_k^r \ge 2, and \; q_k^r\in\mathbb{Z}^{+},\forall k\in \mathcal{K}^r, \tag{\ref{eq:problem}a}\\
	&\nonumber 1 \leq K \leq N,\tag{\ref{eq:problem}b}
\end{align}
where $\omega_{1}$ and $\omega_{2}$ are the nonnegative weighted factors.

The optimization problem \textbf{P1} is challenging to solve using conventional methods because it is a non-linear constrained integer optimization problem. Although some techniques, such as the branch and bound method, can provide approximate solutions, their computational complexities are very high. Moreover, the conventional optimization methods are based on the assumption that the channel state information (CSI) is known. However, it is impractical to obtain all CSI information in the real environment\cite{10032267}. To address these challenges, DRL can be employed to handle imperfect information. By interacting with dynamic channel environments, DRL can train a model by learning from online user behavior and environmental conditions, and thus maximizes the system utility\cite{9575181}. This makes DRL a promising solution for tackling these problems. Since the centralized DRL framework results in significantly high traffic load, and the high mobility of vehicles makes it difficult for the BS to collect accurate CSI. In the next section, we will propose a distributed DRL scheme to solve the optimization problem \textbf{P1}.

\section{Distributed Deep Reinforcement Learning Based Solution}
\label{sec6}

We model the quantization level allocation problem in each round as a distributed DRL process. The duration of each round consists of multiple time steps. For simplicity, we omit the label of round in the DRL process since each round will experience the same training process. In each time step $t$ of a round, each selected vehicle $V_k$ observes the current local state $s_{k,t}$ and takes an action $a_{k,t}$ according to the policy $\pi$. Then, the vehicle receives the reward $r_{k,t}$ and the environment state changes from $s_{k,t}$ to $s_{k,t+1}$ to start the process in the next time step. Next, the DRL framework including state $s_{k,t}$, action $a_{k,t}$ and reward $r_{k,t}$ of the vehicle at time step $t$ will first be constructed, then the DRL based solution is described.

\subsection{DRL framework }

\subsubsection{State}
Each vehicle $V_k$ observes its local state to determine the quantization level.
In the system model, each vehicle calculates the distance between $V_k$ and BS at time step $t$, i.e., $d_{k,t}$, to determine the position of $V_k$ at time step $t$, which reflects the mobility of $V_k$, thus $d_{k,t}$ is set as one local observation at time step $t$. In addition, the BS detects the SNR of $V_k$ at time step $t$, i.e., $\gamma_{k,t}$, and transmits $\gamma_{k,t}$ to $V_k$ at the next time step. In this case, $V_k$ receives $\gamma_{k,t-1}$ at time step $t$, which can reflect the uncertain channel condition at time step $t$, thus $\gamma_{k,t-1}$ is set as the other local observation at time step $t$. Moreover, according to Eqs. \eqref{eq:S}, \eqref{eq:T_epsilon} and \eqref{eq:E_n}, the optimization objective is affected by the number of bits for the transmission, the minimum convergence round and the quantization error, all the three factors are the functions of the quantization level $q_{k,t}$, thus $q_{k,t}$ is set as another local observation at time step $t$. According the above analysis, the state of $V_k$ at time step $t$, denoted as $s_{k,t} \in \mathcal{S}$, where $\mathcal{S}$ is the state space, can be defined as

\begin{equation}
s_{k,t}=[\gamma_{k,t-1},d_{k,t},q_{k,t}].
\label{eq18}
\end{equation}

\subsubsection{Action}
Since each vehicle $V_k$ selects its quantization level to improve FL performance, the action of $V_k$ at time step $t$ is defined as
\begin{equation}
a_{k,t}=q_{k,t}.
\label{eq:a_nt}
\end{equation}
where $q_{k,t} \in \mathcal{A}$ and the space of the quantization level $\mathcal{A}$ is partitioned into $\mathcal{A}=[2,\dots,10]$.
\subsubsection{Reward function}
Since $V_k$ aims to improve FL performance in terms of the training total delay and quantization error, the reward function of $V_k$ at time step $t$ is defined as
\begin{equation}
r_{k,t}= - \left[\omega_{1}\cdot (R_\lambda\cdot T_k^{r,fed})+\omega_{2}\cdot E_k^r\right].
\label{eq:r_nt}
\end{equation}

The expected long-term discounted reward of $V_k$ is calculated as
\begin{equation}
G(\pi_k) \coloneqq \mathbb{E}_{\pi_k}\left[\sum_{t=1}^{T}\gamma^{t-1}r_{k,t}\right],
\label{eq:G}
\end{equation}
where $\gamma \in [0,1]$ is the discount factor and $T$ is the upper limit of step index. The objective of this paper is to find the optimal policy $\pi_k^*$ to maximize the expected long-term discounted reward of $V_k$.


\subsection{Solution}

In this section, we first describe the training stage to obtain the optimal policy, then introduce the testing stage to test the performance under the optimal policy.

\subsubsection{Training stage}

Since the state and action space are discrete and the DDQN algorithm is suitable to solve the DRL problem under the discrete state and action space. Moreover, DDQN can easily find the optimal action when the state-action space $\mathcal{S}\times\mathcal{A}$ is large. Therefore, we utilize the DDQN algorithm to obtain the optimal policy in the training stage.

The DDQN framework includes the target network and the double Q-learning method as an extension of DQN \cite{Lee2022}. It uses a deep neural network (DNN) to approximate the Q-function that maps state-action pairs $(s_k,a_k)$ to the Q-values $Q(s_k,a_k)$. The Q-function can be approximated by using the weight parameter $\theta_k$.

A DNN model consists of an input layer, a hidden layer, and an output layer, where each layer is fully connected to the next layer. The input of the DNN is the state vector $s_{k,t}$ and the output is the Q-value corresponding to each action. The output layer consists of linear units, and the intermediate hidden layer is activated by rectified Linear units (ReLUs) $f(x)=\text{max}(0,x)$.


DDQN decouples the target network $\theta_k^{'}$ as well as the prediction network $\theta_k$ to reduce the correlation between the prediction value and target value. The prediction value $Q(s_{k,t},a_{k,t};\theta_k)$ is obtained from the prediction network $\theta_k$ and the target value $y_{k,t}$ is obtained from the target network $\theta_k^{'}$.

Here, the prediction network $Q(s_k,a_k;\theta_k)$ is used for action selection and the target network $Q(s_k,a_k;\theta_k^{'})$ is used for value evaluation. The pseudocode of the DDQN training process is described in Algorithm \ref{al2}.
\begin{algorithm}
	\caption{Training Stage for the DDQN based Framework}
	\label{al2}
	Initialize replay experience buffert $\mathcal{B}$\\
	Randomly initialize $\theta_k$\\
	Initialize target network by $\theta_k^{'}\leftarrow\theta_k$\\
	\For{episode from $1$ to $T$}
	{
		Reset simulation parameters for the VEC system model\\
		
		Receive initial observation state $s_{k,0}$\\
		\For{time step $t$ from $1$ to $T_I$}
		{
			\If {$p_{\epsilon}\geq \epsilon$}
			{Randomly select action $a_{k,t}\in \mathcal{A}$}
			\Else
			{Select RL-based greedy action $a_{k,t}=arg\;\underset{}{max}Q(s_{k,t},a_{k,t};\theta_{k})$}
			Observe environment and compute next state $s_{t+1}$ and $r_{t+1}$\\
			Store experience $(s_{k,t},a_{k,t},r_{k,t+1},s_{k,t+1})$ into $\mathcal{B}$\\
			\If {number of tuples in $\mathcal{B}\geq I$}
			{
				Randomly sample a mini-batch of $I$ transitions tuples from $\mathcal{B}$\\
				\For{$j$ in $I$}
				{Set target value $y_j$ according to Eq. \eqref{eq:yt1}\\
					Update the predict network by minimizing the loss function}
				\For{every $C$ time steps}
				{Update target network $\theta_k^{'}\leftarrow\theta_k$}
			}
		}
	}
\end{algorithm}


Firstly, the prediction Q-network $\theta_k$ and the target Q-network $\theta_k^{'}$ are initialized, where the prediction network $\theta_k$ is randomly initialized and the target network is initialized as the prediction network $\theta_k$. A replay buffer $\mathcal{B}$ with enough space is constructed to cache the transitions at each time step (lines 1-3).

Then, the algorithm will iteratively train the prediction Q-network and target Q-network in different episodes. In the first episode, the location of $V_n$ is reset to where it enters the coverage area of the BS, i.e., $d_{k,0}$ is set to $-R_B$. Then quantization level $q_{k,0}$ and channel gain $h_{k,0}$ are randomly initialized. As we can calculate the initial SNR according to Eq. (\refeq{eq:SINR_n}), $V_n$ can observe the state at time step $0$, i.e., $s_{k,0} = [\gamma_{k,0}, d_{k,0}, q_{k,0}]$ (Lines 4-6).

Afterwards, in one time step, based on a probability threshold $\epsilon$, an action is randomly selected from the action space $\mathcal{A}$ for exploration. Otherwise, the RL training process selects the greedy action. Mathematically, it can be expressed as  (lines 7-11).
\begin{equation}
	a_{k,t} = \left\{ {\begin{array}{*{20}{c}}
			arg \; \underset{a_k}{\text{max}}Q(s_{k,t},a_k),&{1-\epsilon},\\
			{\text{Randomly choose from}} \;\mathcal{A},&{\epsilon}.
	\end{array}} \right.
	\label{eq:a_t}
\end{equation}

Then, according to the action $a_{k,t}$, the agent can observe the environment and calculate the next state $s_{k,t+1}$ and reward $r_{k,t+1}$ to get the experience. The experience $(s_{k,t}, a_{k,t}, r_{k,t+1}, s_{k,t+1})$ is then stored in the replay buffer (lines 12-13).

When the number of the stored tuples is larger than $I$, the parameters of prediction network and target networks, i.e., $\theta_k$ and $\theta_k^{'}$, are updated literately.

The iteration in time step $t$ ($t=1,2,...,T_I$) to update $\theta_k$ and $\theta_k^{'}$ is described as follows when the number of the stored tuples is larger than $I$. each vehicle first uniformly samples $I$ tuples from replay buffer to form a mini-batch. Then each vehicle inputs each tuple into the target network and prediction network. Each vehicle first inputs $s_{k,t}$ and $a_{k,t}$ into the prediction network and outputs the prediction value $Q(s_{k,t},a_{k,t};\theta_k)$, then each vehicle inputs $s_{k,t+1}$ and $r_{k,t+1}$ into the target network and calculates the target value as
\begin{equation}
	y_{k,t}=r_{k,t+1}+\gamma Q(s_{k,t+1},arg \;\underset{a_k}{\text{max}}Q(s_{k,t+1},a_k;\theta_k);\theta_k^{'}).
	\label{eq:yt1}
\end{equation}

Then the loss function can be calculated as
\begin{equation}
	L(\theta_k)=\mathbb{E}\left[(y_{k,t}-Q(s_{k,t},a_{k,t};\theta_k))^{2}\right],
	\label{eq:L}
\end{equation}
and the prediction network updates its parameters using $\nabla L(\theta_k)$ to
minimize the loss function through gradient descending (lines 14-18).

\begin{figure*}
	\centering
	\includegraphics[width=6in]{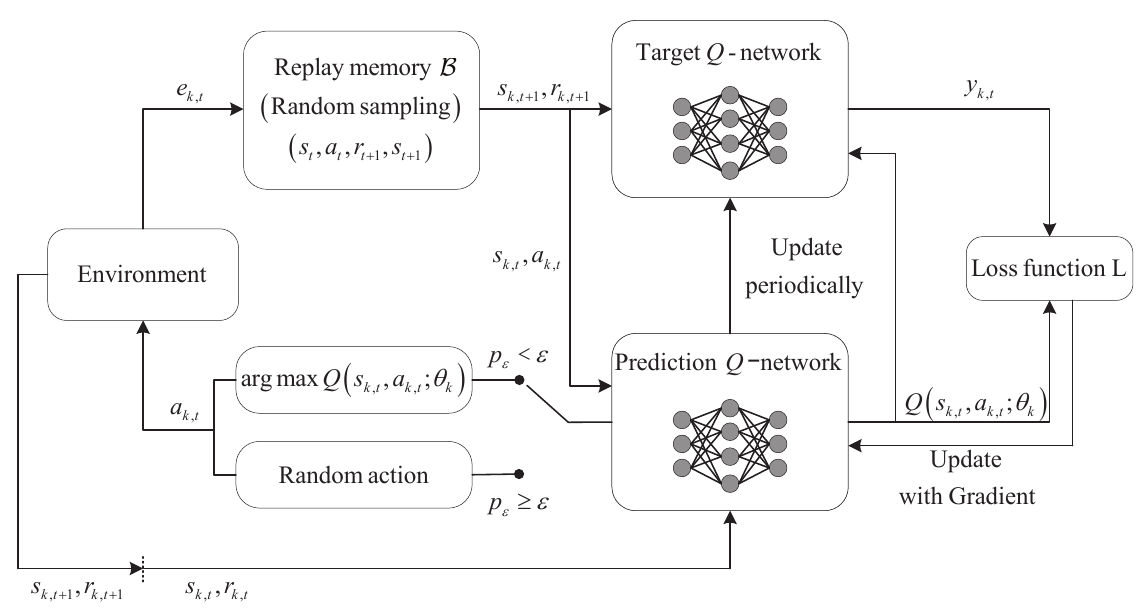}
	\caption{Flow Diagram of DDQN}
	\label{fig2}
\end{figure*}

The target network is updated every $C$ time steps (lines 19-20).
The flow diagram of the DDQN algorithm is shown in Fig. \ref{fig2}.
\begin{algorithm}
	\caption{Testing Stage for the DDQN based Framework}
	\label{al3}
	\For{episode from $1$ to $T'$}
	{
		Reset simulation parameters for the VEC system model\;
		Receive initial observation state $s_{k,0}$\;
		\For{time step $t$ from $1$ to $T_I$}
		{
			Generate the action according to the optimal policy
			$a_{k,t}=arg\;\underset{}{max}Q(s_{k,t},a_{k,t};\theta_k)$\;
			Execute action $a_{k,t}$, observe reward $r_{k,t}$ and new state $s_{k,t+1}$ from the system model.
		}
	}
\end{algorithm}
\subsubsection{Testing stage}
To evaluate the performance of the proposed algorithm, we employ the optimal neural network parameters obtained in the training stage and start to test its performance. The pseudocode of the testing stage is shown in Algorithm \ref{al3}.

\begin{figure}[hb]
	\centering
	\includegraphics[scale=0.46]{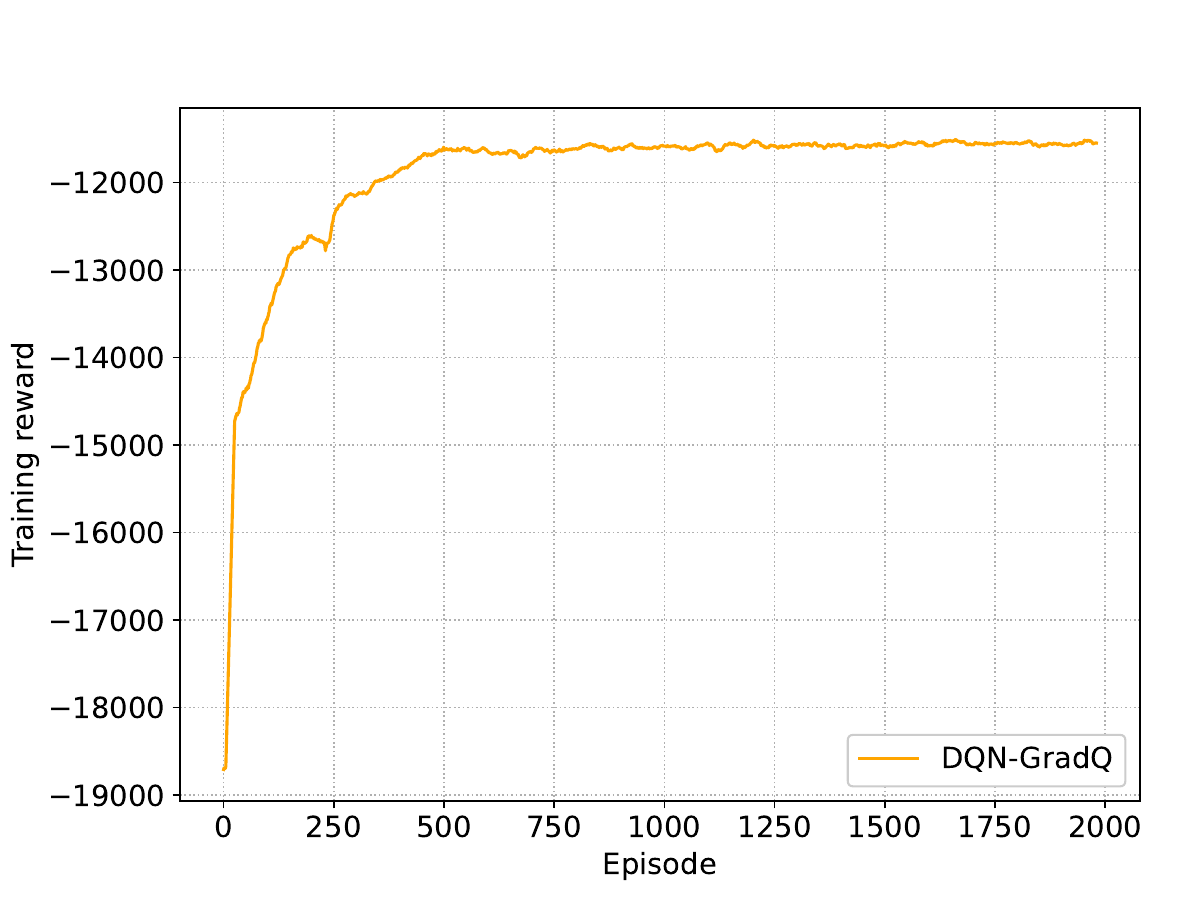}
	\caption{Learning curve.}
	\label{fig3}
\end{figure}

\subsection{Complexity Analysis}

In this subsection, we analyze the complexity of the DDQN algorithm. The complexity of DDQN algorithm is affected by the number of step intervals. For each step interval, the prediction network computes the gradients and updates the parameters, while the target network updates the parameters without calculating the gradients. Since the network structure of the prediction network is the same as that of the target network, the complexity of updating the parameters of the target network is the same as that of the prediction network. Let $G$ be the computational complexity of the prediction network to compute the gradient, and $U$ be the computational complexity of the prediction network to update the parameters. Therefore, the complexity of DDQN algorithm in one time step is $O(G + 2U)$. Moreover, the training processes is not activated until the replay experience buffer is larger than the min-batch, Note that the above algorithm loops over $T$ episodes, each of which includes $T_I$ step steps. Therefore, the complexity of the DDQN algorithm is calculated as $O((T\cdot T_I-I)(G + 2U))$.
\section{Simulation Results}
\label{sec7}
\begin{table}[t]
	\caption{Values of the parameters in the experiments.}
	\label{tab2}
	\footnotesize
	\centering
	\begin{tabular}{|c|c|c|c|}
		\hline
		\multicolumn{4}{|c|}{Parameters of System Model}\\
		\hline
		\textbf{Parameter} &\textbf{Value} &\textbf{Parameter} &\textbf{Value}\\
		\hline
		$\sigma^2$ &$10^{-9}$ W &$p_n$ &$23$ dBm\\
		\hline
		$R_B$ &$500$ m& $B$ & $1$ MHz\\
		\hline
		$W$ &$12$ &$v_n$ &$10$ m/s \\
		\hline
		$c$ &$2.5\times10^{10}$ &$f_n$ &$500$ MHz\\
		\hline
		$d$ &$269722$ &$H$ &$10$ m\\
		\hline
		$\alpha$ &$2$ &$\mathcal{N}$ &$15$ \\
		\hline
		\hline
		\multicolumn{4}{|c|}{Parameters of DDQN}\\
		\hline
		\textbf{Parameter} &\textbf{Value} &\textbf{Parameter} &\textbf{Value}\\
		\hline
		$\gamma$ &$0.99$ &$C$ &$1000$\\
		\hline
		$T$ &$2000$ &$I$ &$64$\\
		\hline
		$T^{'}$ &$500$ &$|\mathcal{B}|$ &$2.5\times10^5$\\
		\hline
		$T_I$ &$1000$ &$\epsilon$ &$0.5$\\
		\hline
	\end{tabular}
	\vspace{-8pt}
\end{table}
\begin{figure*}[b]
	\centering
	\vspace{-20pt}
	\subfloat[]{\includegraphics[scale=0.46]{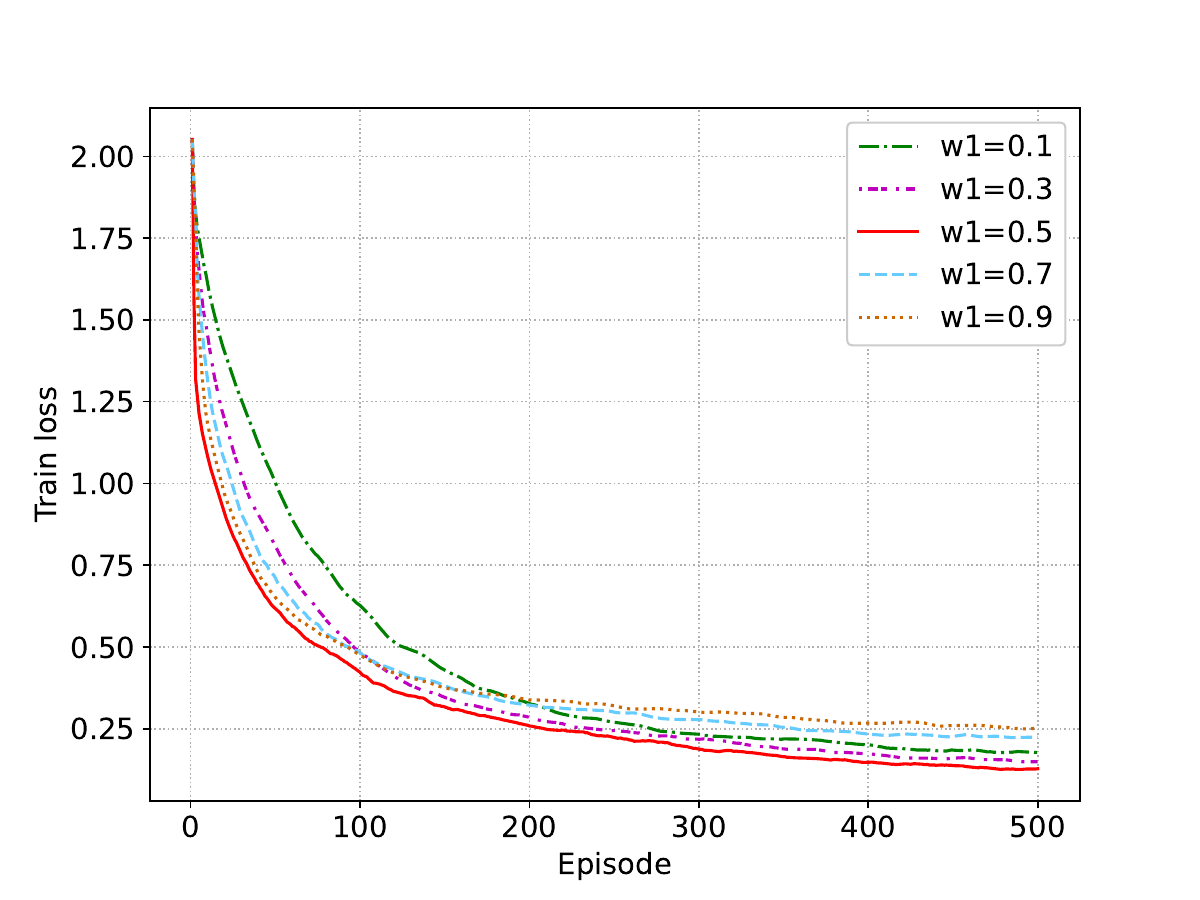}
		\label{fig_train_loss_w}}
	\subfloat[]{\includegraphics[scale=0.46]{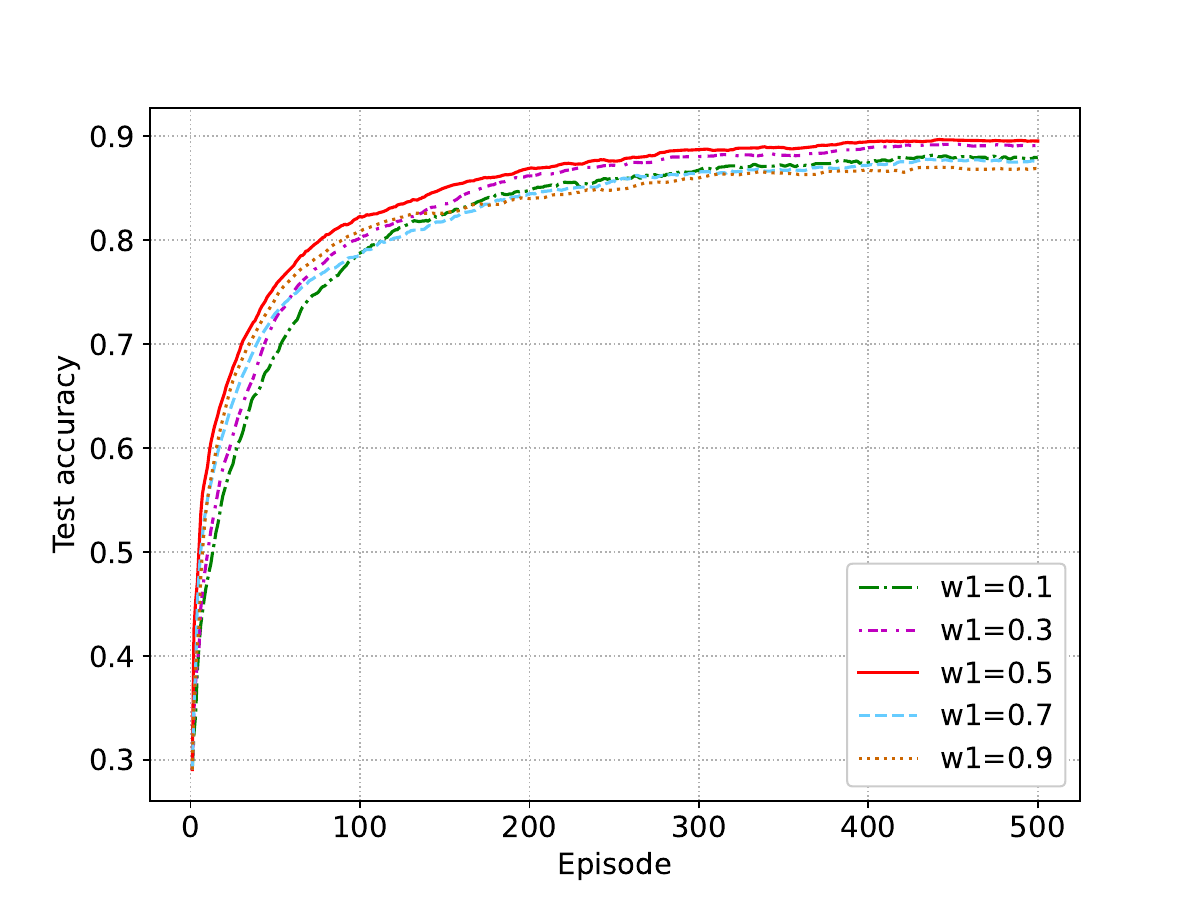}
		\label{fig_test_acc_w}}\\
	\vspace{-10pt}
	\subfloat[]{\includegraphics[scale=0.305]{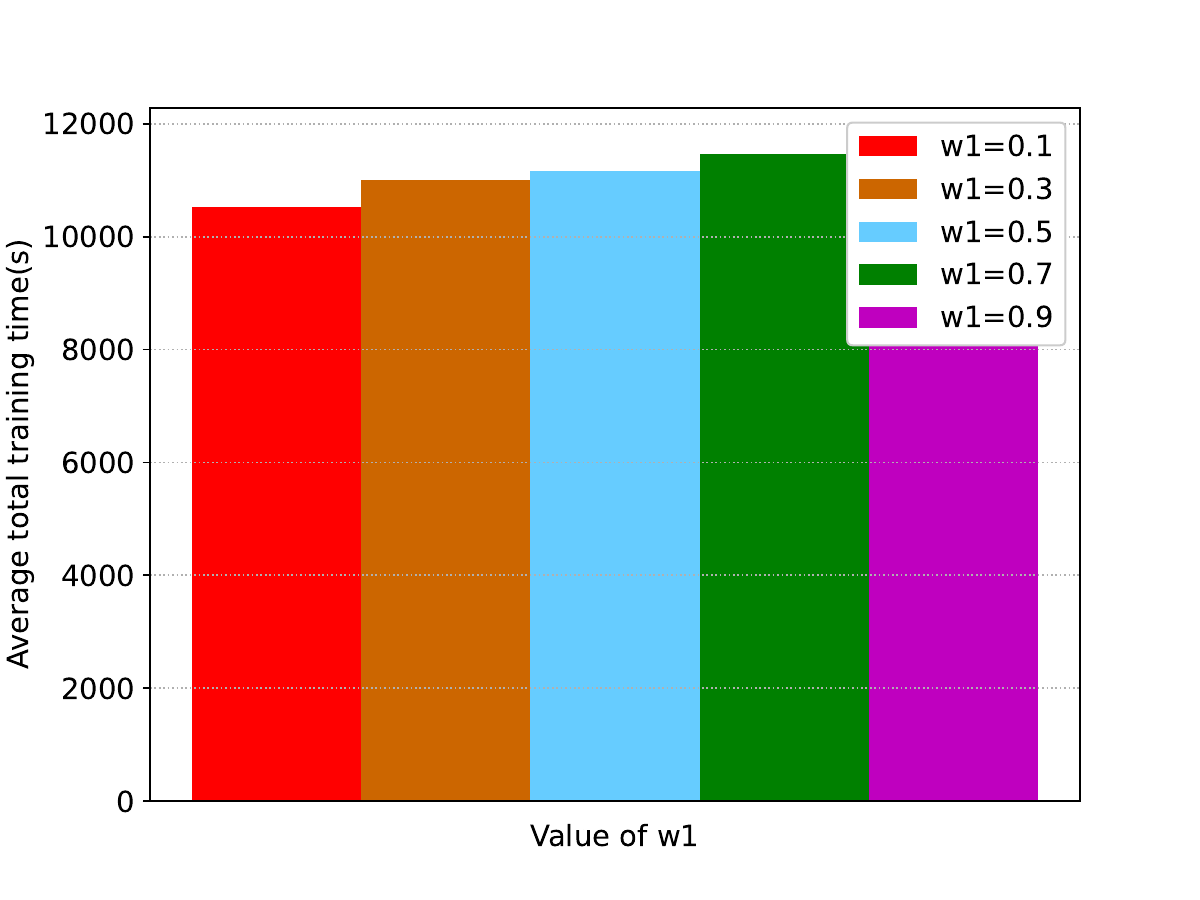}
		\label{fig_time_w}}
	\subfloat[]{\includegraphics[scale=0.305]{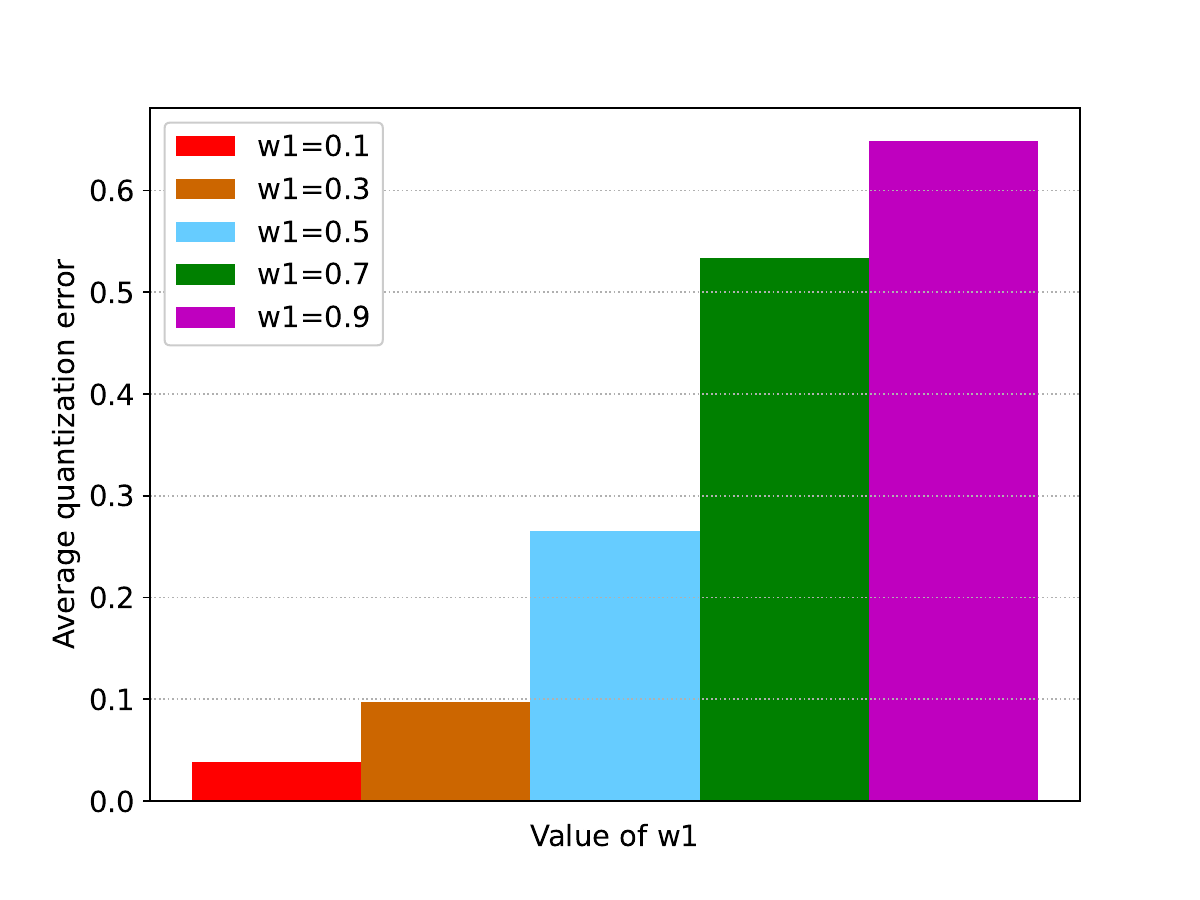}
		\label{fig_qerror_w}}
	\subfloat[]{\includegraphics[scale=0.305]{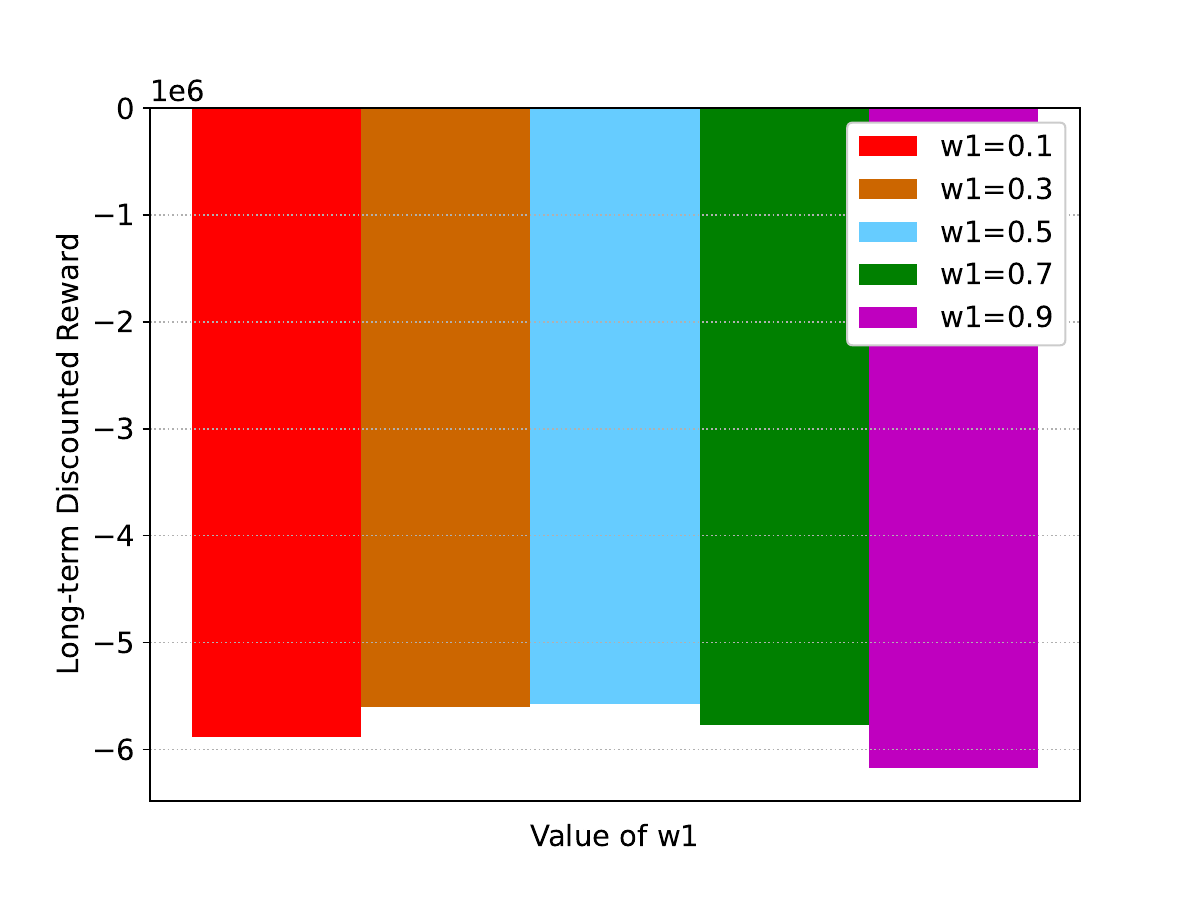}
		\label{fig_reward_w}}\\
	\caption{Performance of FL. (a) Train loss; (b) Test accuracy; (c) Average total training time; (d) Average quantization error; (e) Long-term discounted reward.}
	\label{fig4}
\end{figure*}
In this section, the optimal weight factors between the total training time and QE are figured out through simulation comparisons. On the basis of the optimal weight factors, we conduct simulations to verify the effectiveness of the proposed DDQN-based quantization level allocation algorithm in the training phase and testing phase. All simulations are based on hardware environment (Intel Core i7 CPU) and software environment (Python 3.8.6). We consider the VEC network proposed in Section III, including a BS and several vehicles. Each participating vehicle is treated as an agent and participates in the DDQN-based quantization level allocation algorithm. Excluding $\omega_{1}$ and $\omega_{2}$, the simulation parameters are shown in Table \ref{tab2}.

\subsection{Training Stage}
In Fig. \ref{fig3}, taking the weight setting of $\omega_{1}=\omega_{2}=0.5$ as representative, we investigate the relationship between the average reward per training episode and the number of training iterations to investigate the convergence behavior of the proposed method, which is referred to as DQN-based gradient quantization method (DQN-GradQ) in the simulations for simplicity. It can be seen that the training reward gradually stabilizes as the number of episodes increases, which verifies the good convergence performance of our algorithm.

\subsection{Optimal weight factors}

The condition $\omega_{1}+\omega_{2}=1$ is set for convenience, allowing the value of $\omega_{2}$ to change in response to changes in $\omega_{1}$. Fig. \ref{fig3} shows the performance of FL in terms of training loss, test accuracy, average total training time, average quantization error, and long-term discounted reward for different values of $\omega_{1}$.

As shown in Fig. 4-(a), when $\omega_{1}=0.1$ and $\omega_{1}=0.3$, the train loss value decreases more slowly at the beginning of the training. This is because the vehicles are farther from the base station initially, resulting in poor channel conditions and larger transmission delays. However, the vehicles tend to select higher quantization levels under these weight settings to reduce quantization error. As the channel conditions improve later on, the impact of quantization error becomes significant, and focusing relatively more on quantization error can lead to a lower train loss value.

$\omega_{1}=0.7$ and $\omega_{1}=0.9$ mean that more attention is given to transmission delay of the vehicles. The training loss decreases rapidly at the early stage of the training due to the large transmission delays. However, due to the relatively less focus on quantization error, lower quantization levels are chosen more often, leading to a higher converged training loss value. When $\omega_{1}=0.5$, equal weight is given to the total training time and quantization error, which helps ensure that the system maintains the quality of model convergence while reducing transmission overhead, resulting in the fastest convergence speed compared to other $\omega_{1}$ values. Similarly, as shown in Fig. 4-(b), the model trained with $\omega_{1}=0.5$ outperforms all other schemes in test accuracy.

Fig. 4-(c) and (d) indicate that $\omega_{1}=0.1$ and $\omega_{1}=0.3$ place more emphasis on quantization error, resulting in smaller quantization error and more frequent selection of higher quantization levels during training, allowing for convergence with fewer communication rounds. Conversely, when $\omega_{1}=0.7$ and $\omega_{1}=0.9$, the vehicles tend to choose lower quantization level number, resulting in larger quantization error and an increase in the total training rounds needed for convergence, further leading to an increase in the total training time. The setting of $\omega_{1}=0.5$ does not favor either objective, and under this weight configuration, the system can maintain a low quantization error while achieving faster convergence.

Fig. 4-(e) compares the long-term discounted reward of the scheme for different values of $\omega_{1}$. The setting of $\omega_{1}=0.5$ effectively balances the total training time and quantization error, representing a robust learning strategy that maintains stable performance across different environments and avoids extreme scenarios, accumulating more rewards in the long term.

In summary, $\omega_{1}=0.5$ is identified as the optimal choice. In federated learning, rapid convergence and high accuracy are two key objectives. The setting of $\omega_{1}=0.5$ encourages the system to find the best balance between reducing communication rounds and controlling quantization error, allowing the system to achieve faster convergence without sacrificing too much accuracy. Moreover, this robust weight setting provides good environmental adaptability. In the dynamically changing environment of VEC, the system needs to adapt to different communication conditions and vehicle mobility. The setting of $\omega_{1}=0.5$ offers sufficient flexibility, enabling the system to adjust its quantization strategy according to the current environmental state, thus maintaining good performance under various circumstances. Hence, we set $\omega_{1}=0.5$ for the following experiments.

\begin{figure}
	\centering
	\vspace{-20pt}
	\subfloat[]{\includegraphics[scale=0.46]{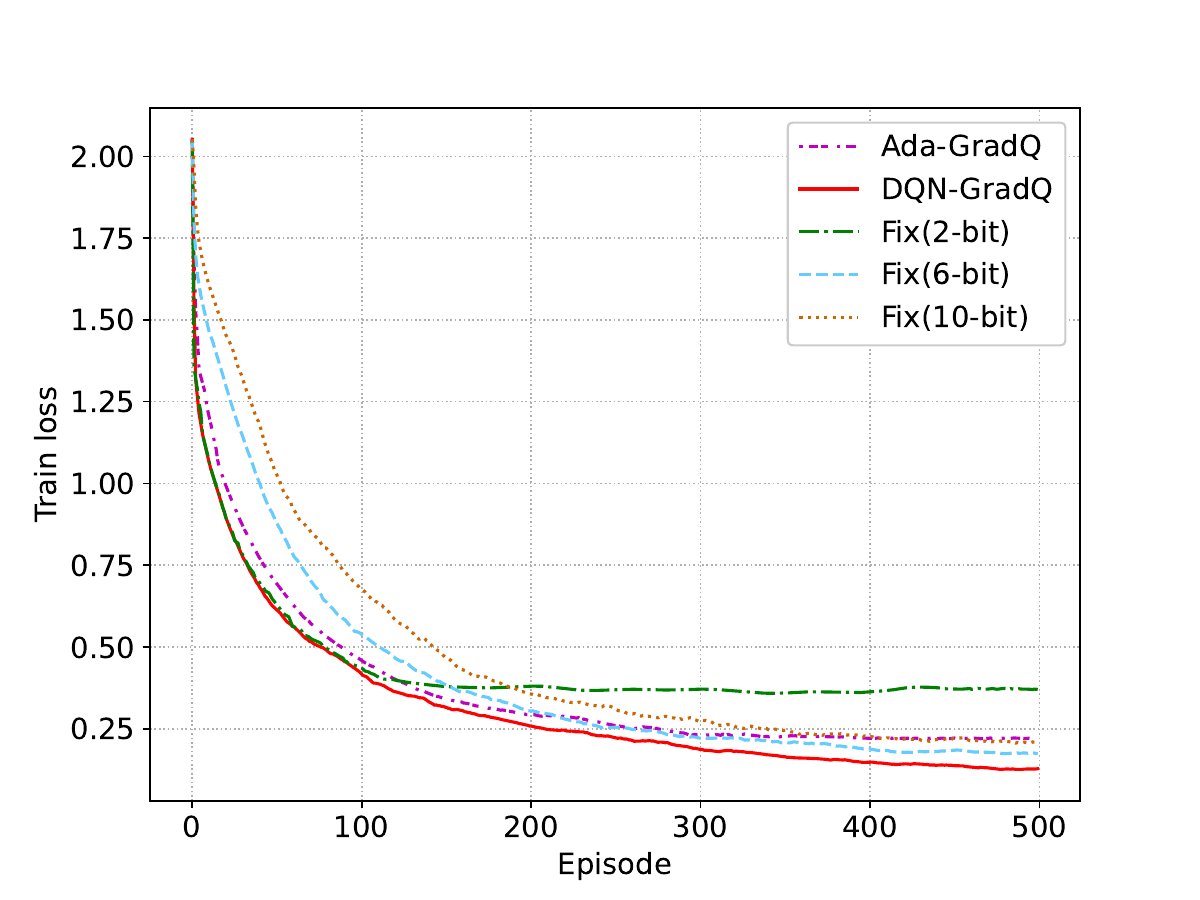}
		\label{fig_train_loss}}\\
	\vspace{-10pt}
	\subfloat[]{\includegraphics[scale=0.46]{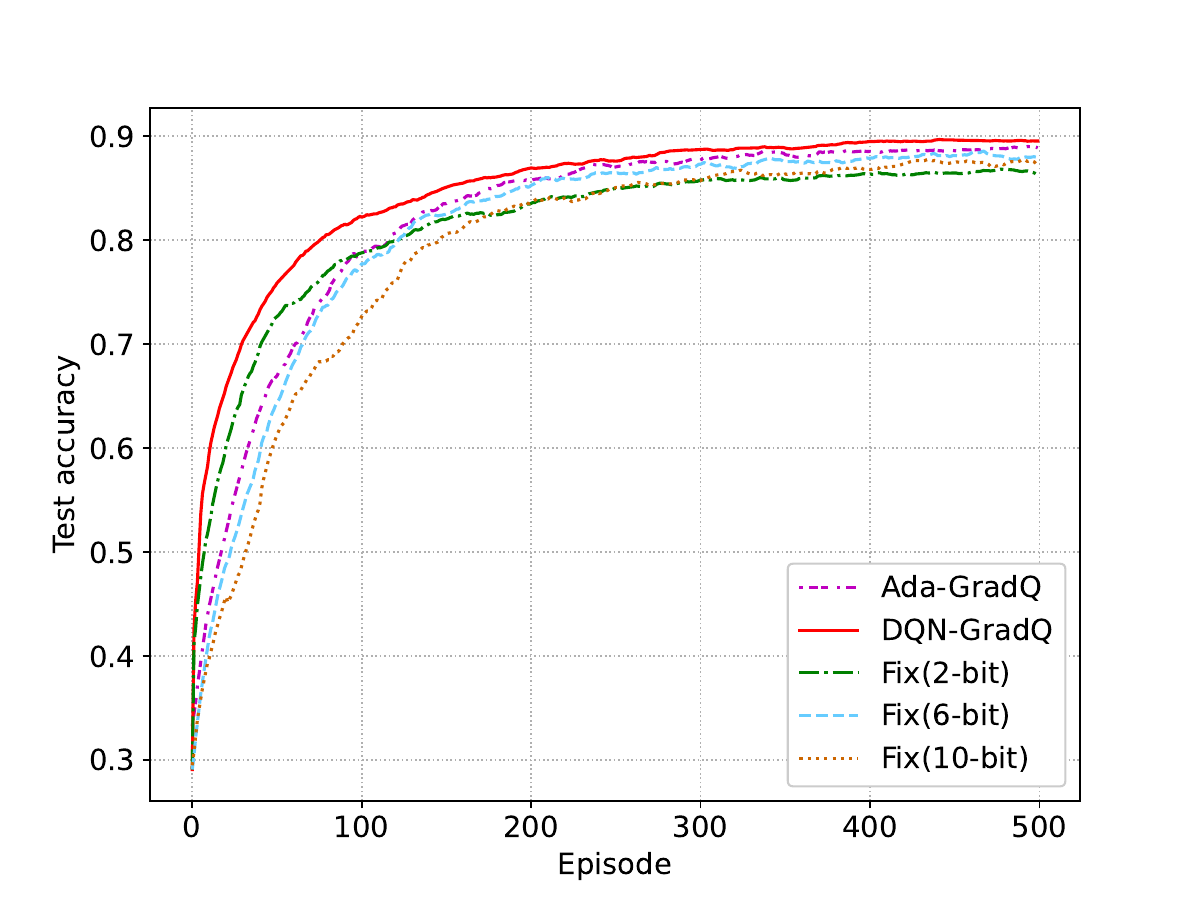}
		\label{fig_test_acc}}\\
	\caption{Performance of FL. (a) Train loss of FL; (b) Test accuracy of FL.}
	\label{fig5}
\end{figure}

\subsection{Testing Stage}

We compare "DQN-GradQ" with several state of the art gradient quantization baselines, including the adaptive quantization method \cite{land2015}, which is referred to as "Ada-GradQ" in the simulations, and the fixed quantization method. For the fixed-bit quantization method, the number of quantization bits is 2, 6, and 10 are referred to as "Fix(2 bits)", "Fix(6 bits)", and "Fix(10 bits)" in the simulations, respectively.
\begin{figure*}
	\centering
	\vspace{-20pt}
	\subfloat[]{\includegraphics[scale=0.305]{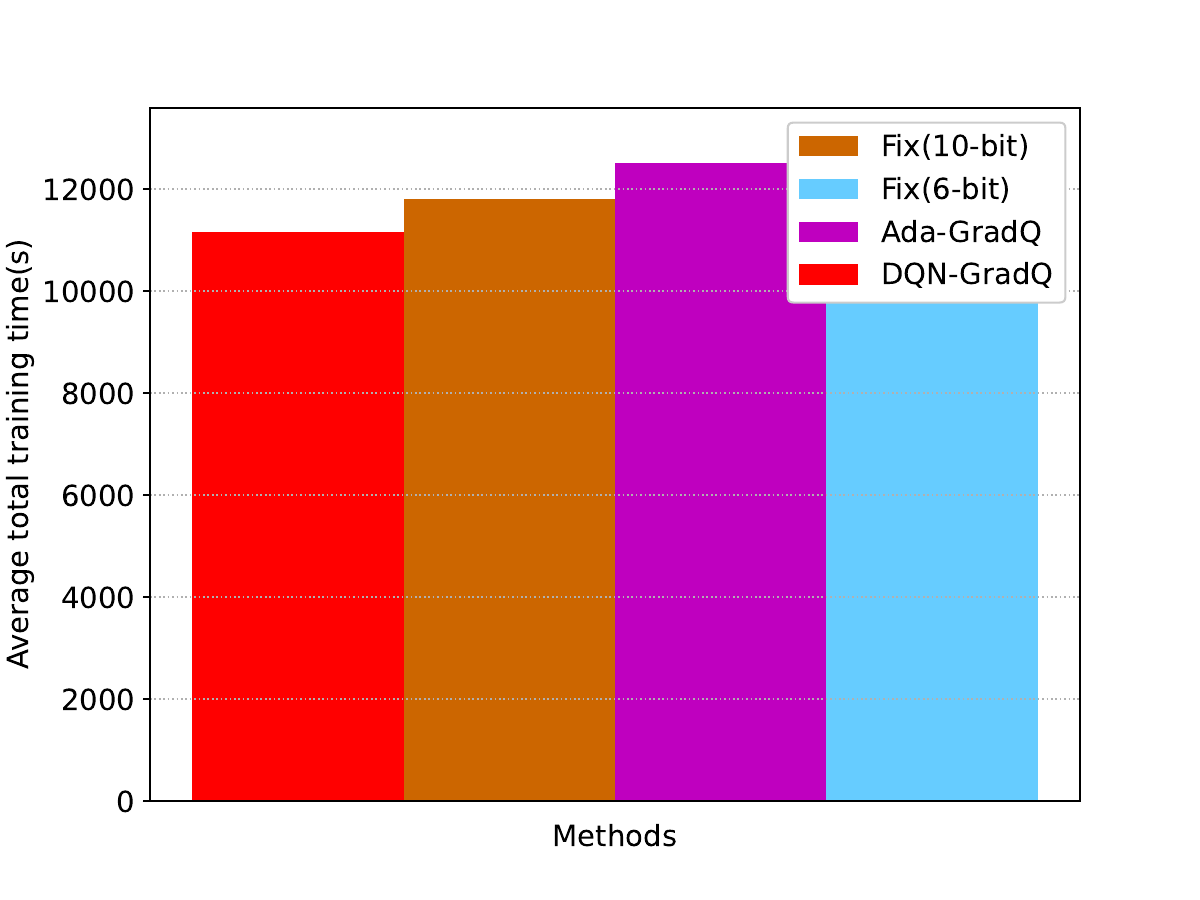}
		\label{fig_avetrainingtime}}
	\subfloat[]{\includegraphics[scale=0.305]{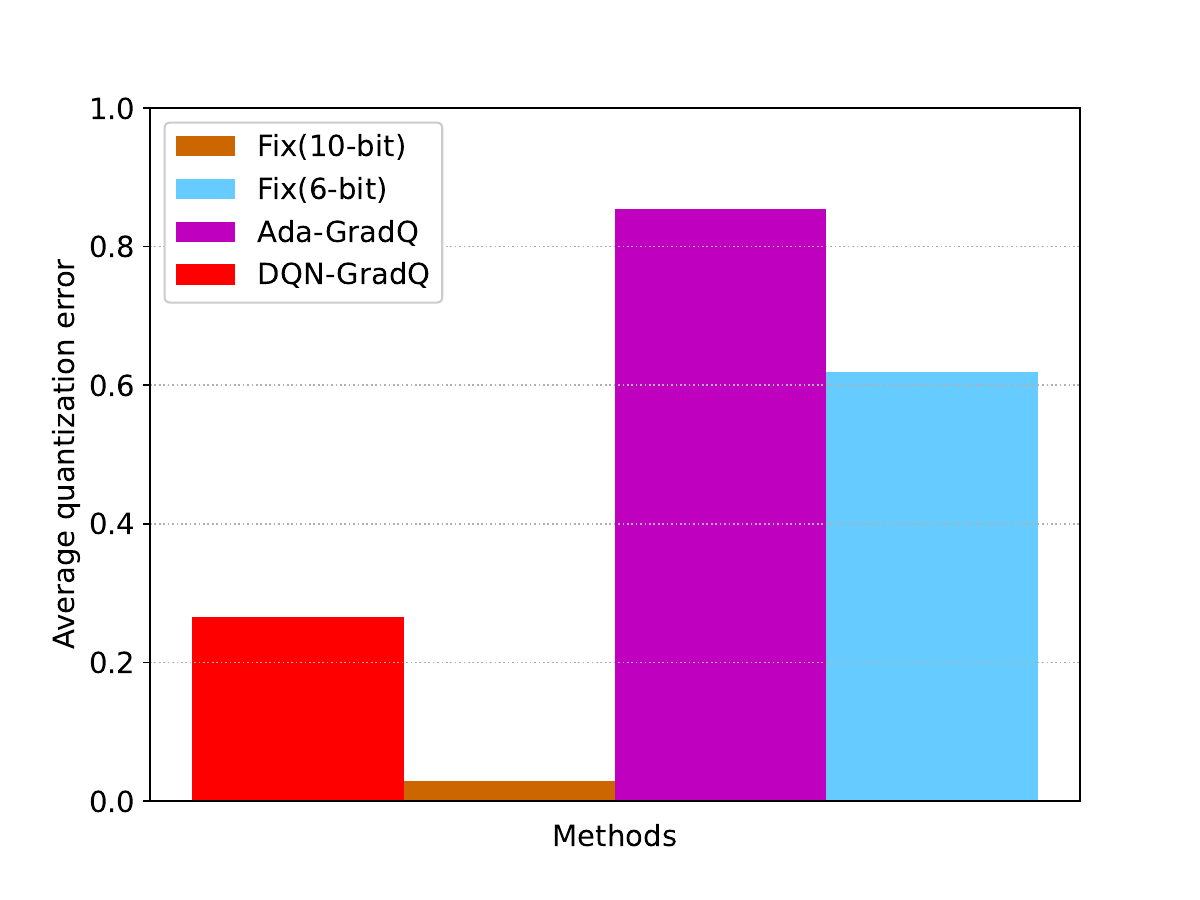}%
		\label{fig_aveqerror}}
	\subfloat[]{\includegraphics[scale=0.305]{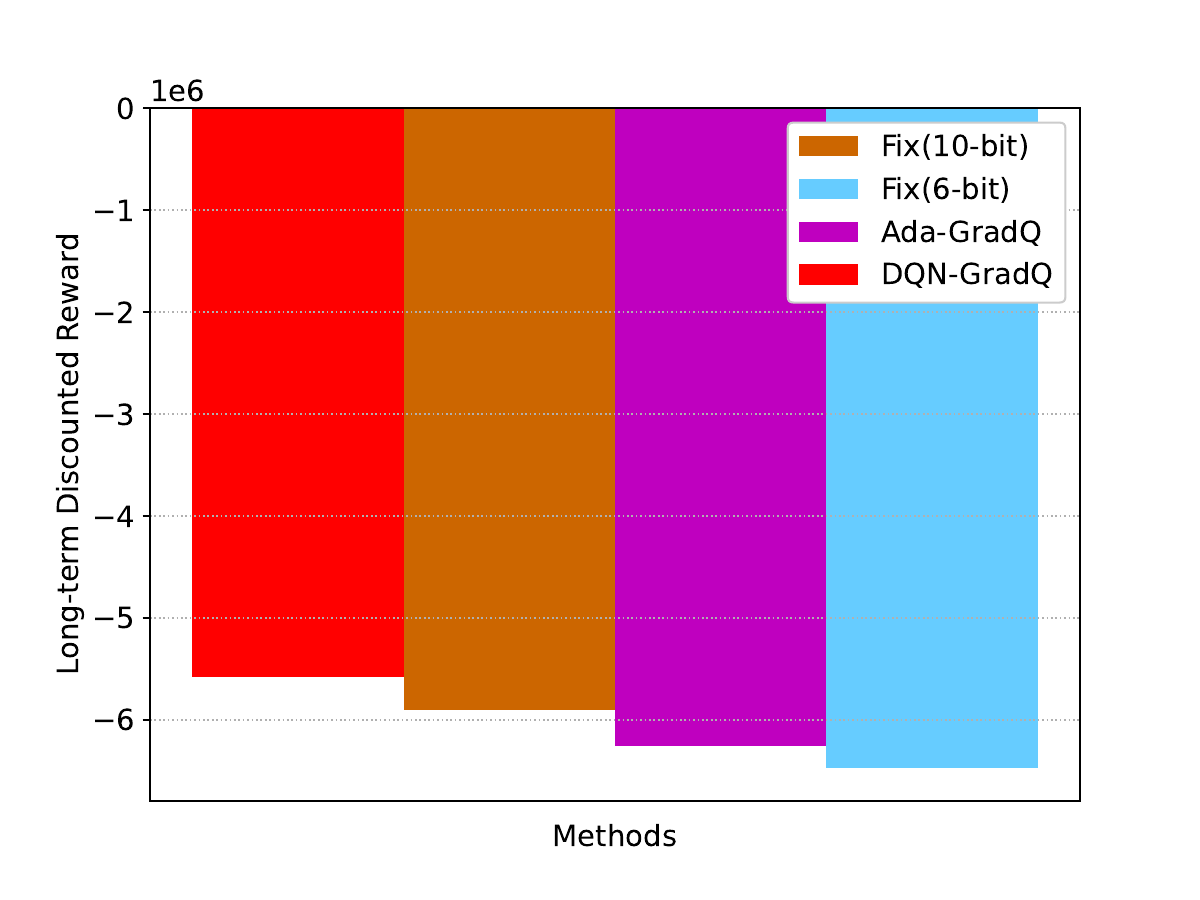}%
		\label{fig_longtermreward}}
    \caption{(a) Average total training time; (b) Average quantization error; (c) Long-term discounted reward. }
	\label{fig6}
\end{figure*}

\begin{figure*}
	\centering
	\vspace{-10pt}
	\subfloat[]{\includegraphics[scale=0.305]{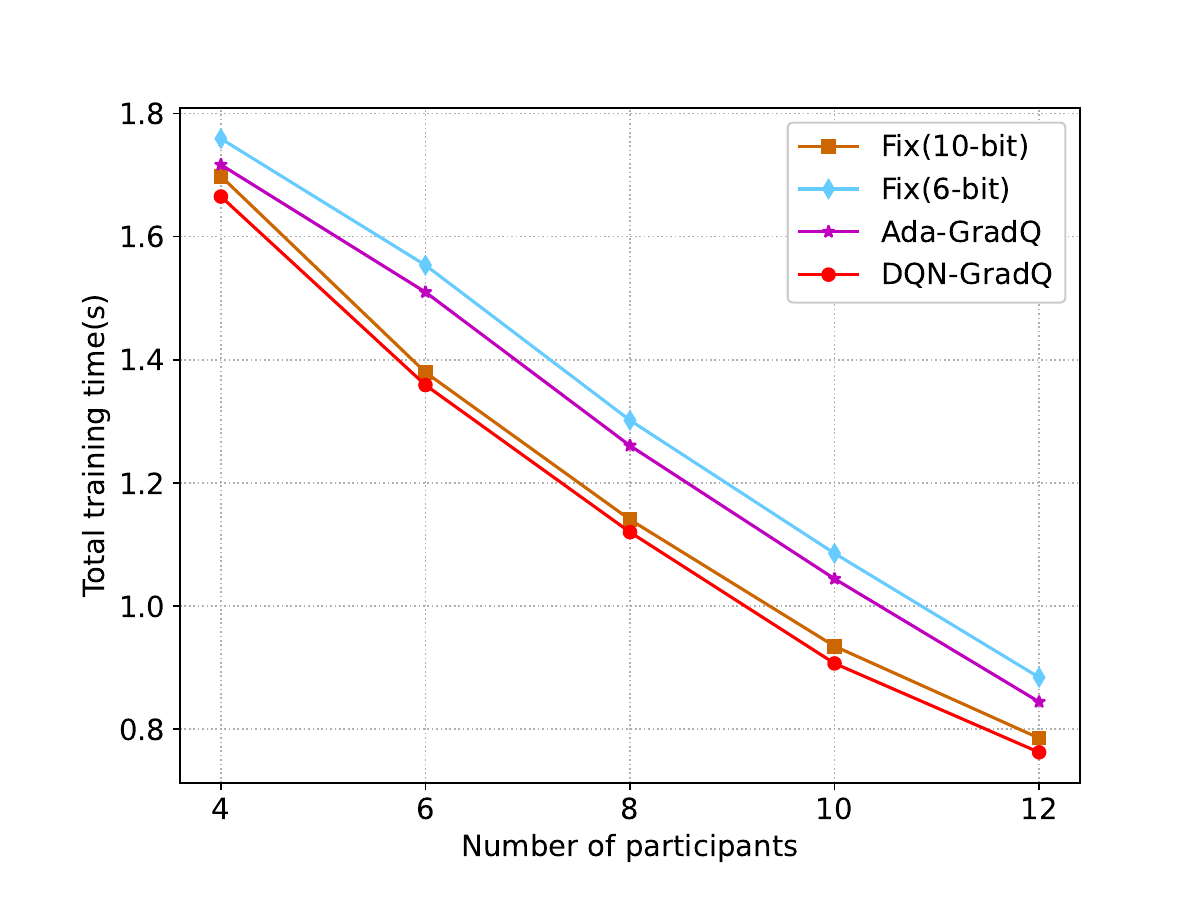}
		\label{fig_total_training_time}}
	\subfloat[]{\includegraphics[scale=0.305]{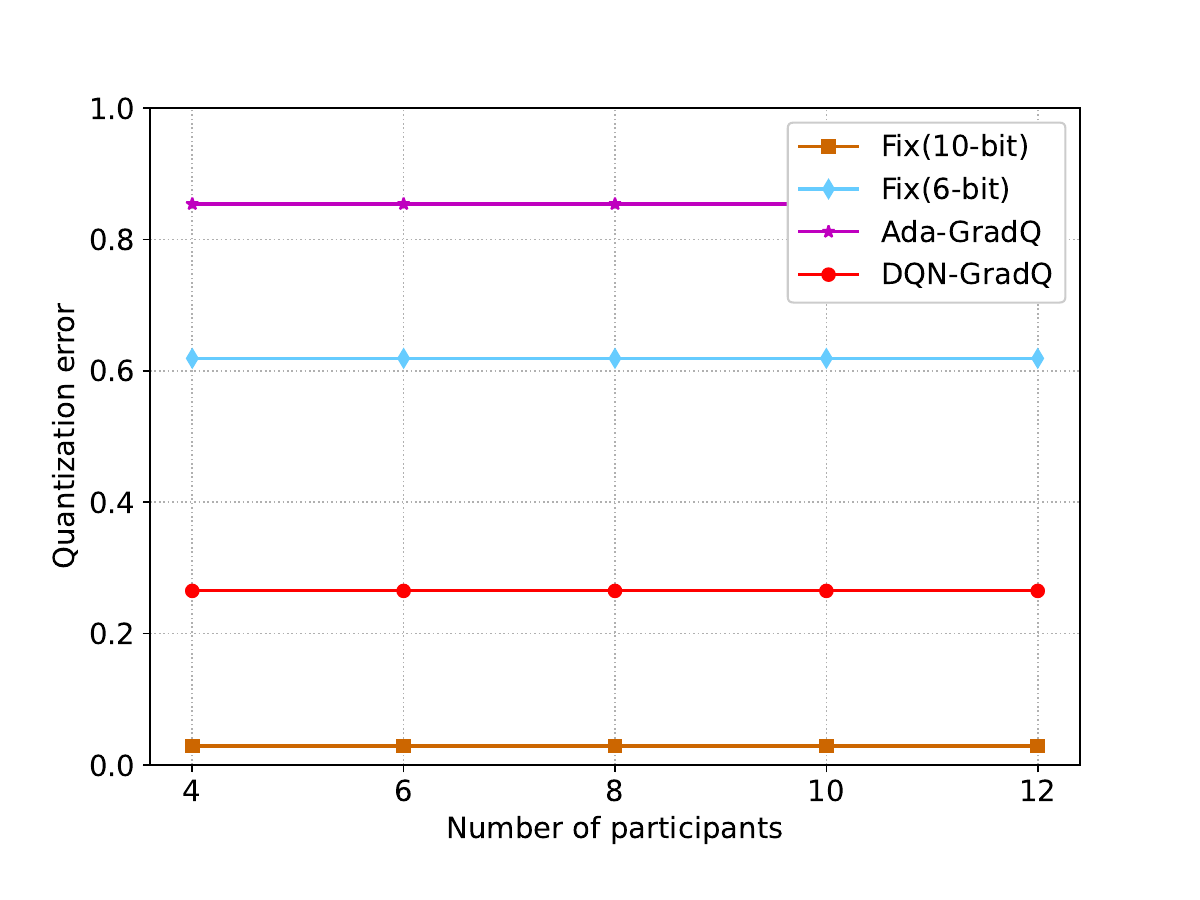}%
		\label{fig_qerrorcompare}}
	\subfloat[]{\includegraphics[scale=0.305]{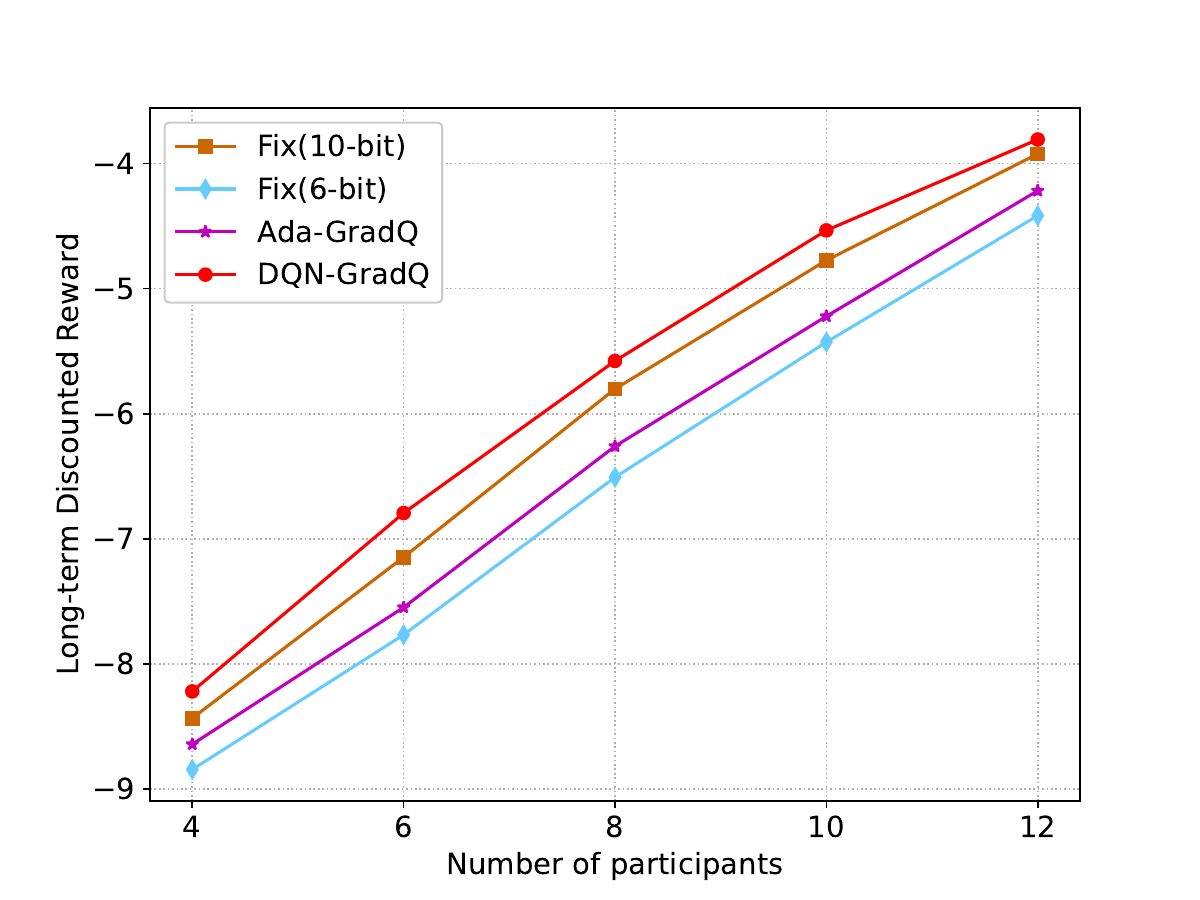}%
		\label{fig_longtermrewardcompare}}
	\caption{Performance vs number of participants. (a) Total training time; (b) Quantization error; (c) Long-term discounted reward.}
	\label{fig7}
\end{figure*}

Fig. 5-(a) and (b) show the training loss and test accuracy of FL under five different quantization schemes, respectively. Fig. 5-(a) shows the training loss comparison for different schemes. We can see that "DQN-GradQ" outperforms all baseline schemes in FL convergence rate. This is because in the early stage of the training, the vehicles are driving far away from the BS, thus the channel conditions are not good owing to the large distances between the vehicles and BS, which incurs the large transmission delay. As the vehicles driving towards the BS, the channel conditions are improve, thus transmission delay of the vehicles are reduced. In this case, the vehicles can gradually increase the quantization level thus reducing the quantization error and converging to the target loss value with less communication rounds. Moreover, we can see that the "Ada-GradQ" approach can get less the number of communication rounds in the early stage of training, but the training loss value is higher than that of "DQN-GradQ". This is because the predefined heuristic algorithm in "Ada-GradQ" cannot make good decisions in the later training phase to guarantee the accuracy of the gradient. In addition, we can see that the training loss of "Fix (2-bit)" decreases very quickly in the early stage, but it converges to a high value. This is because its lower quantization level will produce a larger quantization error. On the other hand, the training curves for "Fix (6bit)" and "Fix (10bit)" steadily decrease throughout the training rounds, however, they need more rounds to converge. This is because they use more bits to quantize the gradient during training and waste a lot of transmission overhead during training, which slows down the convergence speed.
Fig. 5-(b) shows the test accuracy of FL of different schemes. It can be seen that the accuracy of "DQN-GradQ" is higher than that of all baseline schemes under the same training rounds, indicating the lower loss leads to higher performances. Hence, the experimental results show the effectiveness of gradient quantization using DDQN.

Fig. 6 shows the average total training time, average quantization error, and long-term discounted reward of the FL training process under different schemes when the number of the participating vehicle is 8. Fig. 6-(a) shows that our scheme "DQN-GradQ" has the lowest average total training time compared to all the baseline schemes. This is because, the proposed scheme can adaptively adjust the quantization level according to the channel conditions, reduce the number of rounds required for convergence, and then reduce the total training time. Fig. 6-(b) shows that the average quantization error of our scheme is lower than that of "Ada-GradQ" and "Fix (6-bit)" and higher than that of "Fix (10-bit)". This is because the proposed scheme can increase the quantization level and reduce the quantization error according to the improvement of channel conditions. However, Fig. 5-(a) shows that "Fix (10-bit)" cannot converge to a lower loss. Therefore, our scheme can minimize the average total training time of FL while ensuring the convergence of FL compared to all baseline schemes. Fig. 6-(c) compares the long-term discount rewards under different gradient quantization schemes. As can be seen, the long-term discount bonus of "DQN-GradQ" is always higher than that of other schemes. This is because "DQN-GradQ" can adaptively adjust quantization level allocation according to environmental changes to maximize long-term discount rewards.

Fig. 7 shows the total training time, quantization error, and long-term discount reward for the four schemes with different numbers of selected vehicles, respectively. It can be seen from Fig. 7-(a) that the total FL training time of the four schemes gradually decreases with the increase of the number of participating vehicles. Our scheme "DQN-GradQ" is with the considerable lower FL total training time, because more participating vehicles lead to less rounds for convergence, and our scheme dynamically adjusts the allocation of quantization levels according to the change of the environment, hence, the number of communication rounds required for FL convergence is further reduced. It can be seen from Fig. 7-(b) that for the four schemes, the quantization error of different numbers of participating vehicles remains stationary, and our scheme has a relatively low quantization error. This is because the magnitude of the quantization error depends only on the size of the assigned quantization level. That is, the larger the quantization level, the smaller the quantization error, and vice versa. As shown in Fig. 7-(c), with the increase of the number of participating vehicles, the long-term discount rewards of the four schemes gradually increase, and our scheme DQN-GradQ has the largest long-term discount reward. This is because when the number of participating vehicles is larger, the rounds required for convergence is smaller, and the total training time is smaller. The proposed scheme can reasonably allocate the quantization level according to the change of channel conditions to further reduce the training time, which reflects the effectiveness of our scheme in minimizing the training time of FL.

\section{Conclusions}
\label{sec8}
In this paper, we investigated the optimization of the training time and quantization error of FL with gradient quantization in VEC. First, we proposed a mobility and model-aware vehicle selection rule to select vehicles at the beginning of each round. After determining the participating vehicles, we considered the uncertainty of channel conditions caused by vehicle mobility in VEC, and proposed a distributed quantization level allocation scheme based on DDQN to jointly optimize the training time and quantization error of FL. Simulation results demonstrated that the proposed DDQN-based gradient quantization scheme outperforms all benchmark schemes in terms of FL training time, and can minimize the training time while ensuring the FL convergence.

\ifCLASSOPTIONcaptionsoff
  \newpage
\fi

\bibliographystyle{IEEEtran}
\bibliography{IEEEabrv,mybibfile}

\end{document}